\documentclass{article}

    \usepackage[final]{neurips_2025}

\usepackage[utf8]{inputenc} % allow utf-8 input
\usepackage[T1]{fontenc}    % use 8-bit T1 fonts
\usepackage[
    pagebackref,
    breaklinks,
    citecolor=uclablue,
    colorlinks=true,
]{hyperref}
\usepackage{url}            % simple URL typesetting
\usepackage{booktabs}       % professional-quality tables
\usepackage{amsfonts}       % blackboard math symbols
\usepackage{nicefrac}       % compact symbols for 1/2, etc.
\usepackage{microtype}      % microtypography
\usepackage{xcolor}         % colors

\usepackage{marvosym}

\usepackage{url}
\usepackage{multirow} 
\usepackage{graphicx}
\usepackage[export]{adjustbox}
\usepackage{float}
\usepackage{epsfig}
\usepackage{graphicx}
\usepackage{amsmath}
\usepackage{amssymb} % http://ctan.org/pkg/amssymb
\usepackage{caption}
\usepackage{xparse} % data cards
\usepackage{wrapfig} % wrapped table
\usepackage{tabularx}
\usepackage{subcaption}
\usepackage{fancyhdr}
\usepackage{color}
\usepackage{wrapfig} % wrapped table
\usepackage{framed}
\usepackage{soul}
\usepackage{xspace}
\usepackage{booktabs} % professional-quality tables
\usepackage{makecell}
\usepackage{xcolor}
\usepackage{vcell}
\usepackage{colortbl}
\usepackage[nodisplayskipstretch]{setspace}
\usepackage{seqsplit} 
\usepackage{array}
\usepackage{todonotes}
\usepackage{CJKutf8} % Load the CJKutf8 package(Chinese)
\usepackage[normalem]{ulem}
\usepackage{enumitem}
\usepackage{etoc}
\usepackage{tcolorbox}

\newcommand{\header}[1]{\text{#1}}

\definecolor{backred}{RGB}{255, 190, 190}
\definecolor{backblue}{RGB}{210, 230, 250}
\definecolor{myblue}{RGB}{6, 174, 226}

\definecolor{darkgreen}{rgb}{0.0,0.5,0.0}

\definecolor{shadecolor}{RGB}{237,237,237}

\definecolor{uclablue}{rgb}{0.15, 0.45, 0.68}
\definecolor{Gray}{gray}{0.93}
\definecolor{uclagold}{rgb}{1.0, 0.7, 0.0}
\definecolor{airforceblue}{rgb}{0.36, 0.54, 0.66}
\definecolor{rosegold}{rgb}{0.72, 0.43, 0.47}
\definecolor{pastelbrown}{rgb}{0.51, 0.41, 0.33}
\definecolor{isabelline}{rgb}{0.96, 0.94, 0.93}
\definecolor{macaroniandcheese}{rgb}{0.98, 0.89, 0.83}
\definecolor{wildblueyonder}{rgb}{0.85, 0.89, 0.95}
\definecolor{mediumtaupe}{rgb}{0.4, 0.3, 0.28}
\definecolor{bluegray}{rgb}{0.4, 0.6, 0.8}
\definecolor{celestialblue}{rgb}{0.29, 0.59, 0.82}
\definecolor{darkorange}{rgb}{1.0, 0.55, 0.0}
\definecolor{cadmiumred}{rgb}{0.89, 0.0, 0.13}
\definecolor{magnolia}{rgb}{0.97, 0.96, 1.0}
\definecolor{pastelblue}{rgb}{0.68, 0.78, 0.81}
\definecolor{persiangreen}{rgb}{0.0, 0.65, 0.58}
\definecolor{steelblue}{rgb}{0.27, 0.51, 0.71}
\definecolor{bluebell}{rgb}{0.64, 0.64, 0.82}
\definecolor{dimgray}{rgb}{0.41, 0.41, 0.41}
\definecolor{splashedwhite}{rgb}{1.0, 0.99, 1.0}
\definecolor{lavendergray}{rgb}{0.77, 0.76, 0.82}
\definecolor{lightgray}{rgb}{0.83, 0.83, 0.83}
\definecolor{lavendermist}{rgb}{0.9, 0.9, 0.98}
\definecolor{lightgreen}{HTML}{f8fcf4}
\definecolor{lightblue}{HTML}{dfebf7}
\definecolor{zeroshot}{rgb}{0.9, 0.9, 0.9}
\definecolor{fourshot}{rgb}{0.8, 0.9, 0.8}
\definecolor{eightshot}{rgb}{0.8, 0.8, 0.9}
\definecolor{sixteenshot}{rgb}{0.9, 0.8, 0.8}
\usetikzlibrary{patterns}
\definecolor{blue-violet}{rgb}{0.54, 0.17, 0.89}
\definecolor{coral}{HTML}{FF7F50}

\newcommand{\greencheck}{{\color{green}\checkmark}}
\newcommand{\redx}{{\color{red}$\times$}}
\newcommand{\benchmark}{\textsc{3DMem-Bench}\xspace} %%todo change name 
\newcommand{\model}{\textsc{3DLLM-Mem}\xspace} %%todo change name 

\newcommand{\customfootnotetext}[2]{{%
  \renewcommand{\thefootnote}{#1}%
  \footnotetext[0]{#2}}}%

\title{3DLLM-Mem: Long-Term Spatial-Temporal Memory for Embodied 3D Large Language Model}
\author{
 Wenbo Hu$^{1}$\textsuperscript{\Letter} \quad
  Yining Hong$^{1}$ \quad
  Yanjun Wang$^{1}$ \quad
  Leison Gao$^{1}$ \quad
  Zibu Wei$^{1}$ \quad \\ 
  \textbf{Xingcheng Yao}$^{1}$ \quad 
  \textbf{Nanyun Peng$^{1}$} \quad
  \textbf{Yonatan Bitton$^{2}$} \quad
  \textbf{Idan Szpektor$^{2}$} \quad
  \textbf{Kai-Wei Chang$^{1}$} \\ \\ 
  $^1$University of California, Los Angeles, 
  $^2$Google Research \\
      \vspace{-2mm} \\
    \textbf{\url{https://3dllm-mem.github.io}}
    \vspace{-3mm}
}

\begin{document}
% Equal Contribution. $^{*}$ Equal Advising.

\customfootnotetext{}{\textsuperscript{\Letter}
 Contact at whu@cs.ucla.edu. 
  % $^{1}$UCLA \,
  % $^{2}$Google Research \,
  }
\maketitle
% \todo{Yining: I don't like LLM-Mem3D. Maybe I'll change to something else after}
% \wh{todo: Better phrasing embodied actions in the title}

\begin{figure}[h]
  \centering
  % \fbox{\rule[-.5cm]{0cm}{4cm} \rule[-.5cm]{4cm}{0cm}}
\includegraphics[trim=0cm 24.2cm 5.6cm 0cm, clip, width=\linewidth]{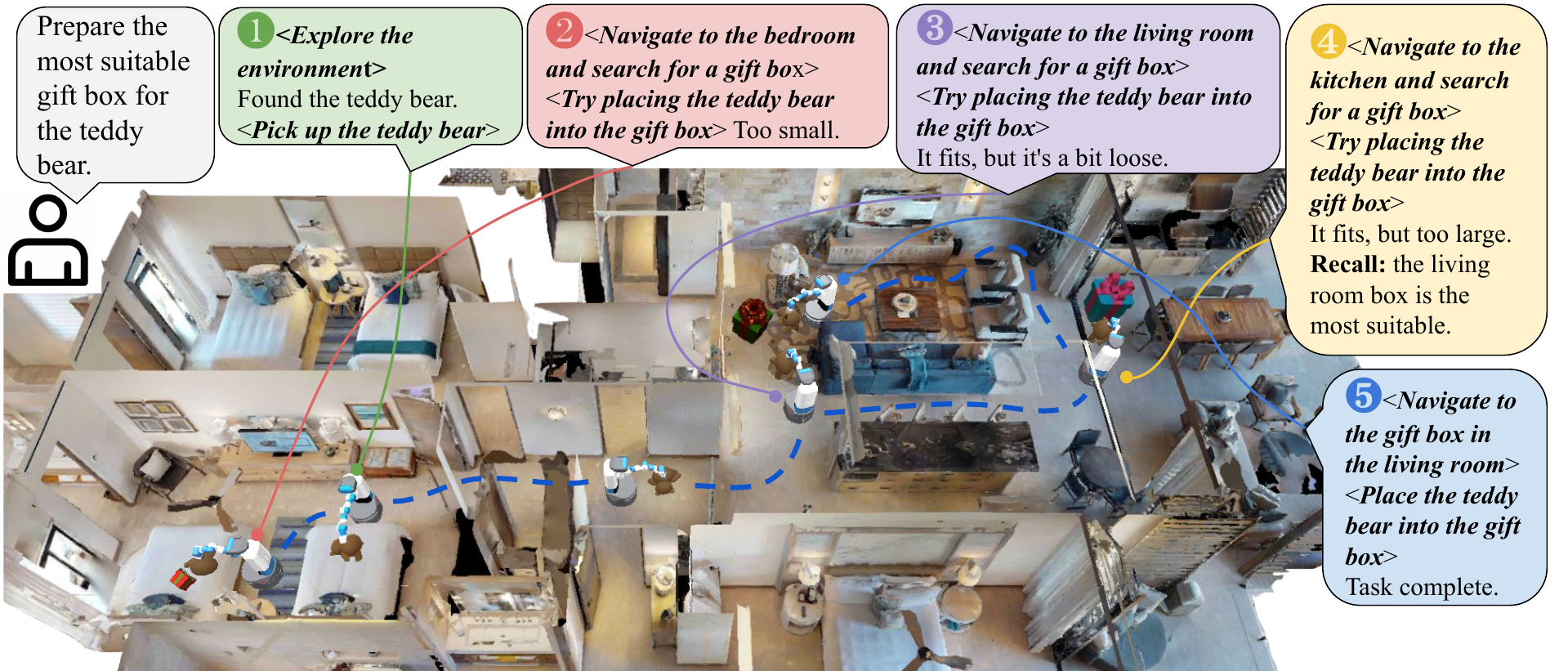}
  \caption{We propose \model, a memory-enhanced 3D embodied agent that explores and incorporates feedback from the environment, interacts with objects, and incrementally builds and maintains a task-relevant long-term memory throughout its trajectory. For illustration purposes, agents from multiple time steps are shown simultaneously.}
 \label{fig:teaser}
\end{figure}

\begin{abstract}

Humans excel at performing complex tasks by leveraging long-term memory across temporal and spatial experiences. In contrast, current Large Language Models (LLMs) struggle to effectively plan and act in dynamic, multi-room 3D environments. 
We posit that part of this limitation is due to the lack of proper 3D spatial-temporal memory modeling in LLMs. 
To address this, we first introduce \benchmark, a comprehensive benchmark comprising over 26,000 trajectories and 2,892 embodied tasks, question-answering and captioning, designed to evaluate an agent's ability to reason over long-term memory in 3D environments.
Second, we propose \model, a novel dynamic memory management and fusion model for embodied spatial-temporal reasoning and actions in LLMs. 
Our model uses \textit{working memory} tokens, which represents current observations, as queries to selectively attend to and fuse the most useful spatial and temporal features from \textit{episodic memory}, which stores past observations and interactions. Our approach allows the agent to focus on task-relevant information while maintaining memory efficiency in complex, long-horizon environments.
Experimental results demonstrate that \model achieves state-of-the-art performance across various tasks, outperforming the strongest baselines by 16.5\% in success rate on \benchmark's most challenging in-the-wild embodied tasks.

\end{abstract}

% (teaser figure, several rooms. task: retrieve the largest teddy bear) Picture yourself situated within within the house in figure xxx and presented with the task... You need to (mention congition, cognitive maps etc. working memory.  episodic memory etc. retrieve the most useful memory from before for current step. chain of actions. plans etc.) Fancier for this paragraph.

% LLMs -> 3D-LLMs -> 3D-VLAs. cannot keep a long-term memory chain. working memory (in the context window) -> episodic memory (out of context window). Some works retrieval of memory. However, for embodied task, spatial memory is also very important. For later, we want to refer to all previous memories. However, if we put all those memories back, the LLMs will be out of memory. And spatial memory and temporal memory are inter-correlated 3D-mem. 

\section{Introduction}
\vspace{-3mm}

Picture yourself traversing an unfamiliar home, as illustrated in Figure~\ref{fig:teaser}, on a mission to explore multiple rooms and evaluate various gift boxes to find the most suitable one for wrapping a teddy bear.
As you navigate from room to room, your brain instinctively creates a 3D cognitive map of the environment, maintains a working memory of objects you've encountered, forms episodic memories that link observations across space and time, and plans efficient actions. 
This seamless integration of 3D spatial understanding, long-term memory encoding and retrieval, fluid switching between working and episodic memory, and purposeful action planning — cognitive processes that humans take for granted — remain formidable challenges for embodied AI systems today.

Recent extensions of Large Language Models (LLMs) to 3D environments have birthed 3D-LLMs~\citep{3dllm, guo2023pointbindpointllmaligning, gu2024conceptgraphs, huang2024embodied, pointllm} that can perceive and reason about 3D spaces, while 3D Vision-Language-Action models~\citep{zhen20243dvla,zhao2025cotvlavisualchainofthoughtreasoning, intelligence2025pi05visionlanguageactionmodelopenworld} further incorporate the ability to plan and act within these environments. Despite these advances, several critical limitations persist that prevent models from performing the kinds of tasks described above. 
First, current models struggle to maintain long-term memory chains when performing complex tasks that unfold across multiple visual scenarios, such as several rooms in a house, and extended time frames. Real-world 3D physical scenes are remarkably vast and information-dense, where every detail can matter for long-horizon embodied tasks — for instance, in Figure~\ref{fig:teaser}, finding the most suitable gift box requires remembering all the gift boxes encountered along the way and their characteristics and interaction with teddy bear. Dense 3D representations are particularly valuable as they capture comprehensive spatial information, preserving intricate geometric relationships and environmental details that sparse or object-centric approaches might miss. However, how to accurately and efficiently store dense 3D memory remains a fundamental challenge - retrieving the entire history would overwhelm the model’s context limits, while selective retrieval~\citep{xie2024embodiedraggeneralnonparametricembodied, wang2025karmaaugmentingembodiedai, yang20243dmem3dscenememory} risks omitting critical information needed for accurate reasoning and decision-making. The second challenge resides in the entanglement of spatial and temporal memory — agents must track not only where objects are, but how they change over time through exploration and interaction. As environments evolve, maintaining coherent representations of previously seen spaces while incorporating new information continues to exceed the capabilities of current embodied AI models.
% First, these models struggle with long-term memory: their limited context windows prevent them from maintaining coherent memory chains over extended spatial-temporal horizons, making it difficult to access past observations crucial for making new decisions. Additionally, they are ill-equipped to generate or manage extended action sequences required to complete complex tasks that unfold across multiple rooms and extended timeframes.
% Moreover,
% for such embodied tasks, spatial memory becomes highly intertwined with temporal memory in terms of tracking how environments and objects change over time through agent explorations and interactions. Effectively integrating relevant past observations with current inputs — without exceeding the model’s memory constraints — remains a core challenge. Retrieving the entire memory history would overwhelm context limits, while selective retrieval \cite{} runs the risk of omitting essential spatial-temporal dependencies necessary for high-level reasoning.

Our efforts at solving this challenge are two-fold. First, we introduce a novel benchmark for reasoning, planning and acting with long-term spatial-temporal memory in embodied environments. Our benchmark, \benchmark, encompasses multi-room 3D scenes from the Habitat environment, augmented with interactive objects to enable manipulation tasks across extended spatial-temporal horizons. 
Notably, we define fine-grained embodied tasks across varying levels of difficulty—from simple to hard—enabling deeper insight into model performance, which we believe is not addressed in prior benchmarks as shown in Table~\ref{table:comparison}. Our task set spans a wide range of complexities, from straightforward object collection to challenging comparative reasoning tasks that require integrating observations across multiple rooms and time steps.
Additionally, we include in-the-wild challenge tasks to evaluate the model's generalization capabilities beyond seen environments.
The benchmark includes three evaluation categories: (1) embodied tasks requiring extended action sequences across multiple rooms, (2) spatial-temporal embodied question answering (EQA) that evaluates understanding of spatial relationships over time, and (3) long-term scene captioning that tests memorization of previously observed environments. 
Our dataset includes 26,000+ trajectory examples 
spanning 182+ unique scenes with an average of 18 rooms per scene.

Second, we introduce \model, a 3D embodied LLM with dynamic memory management capabilities designed specifically for embodied spatial-temporal reasoning, planning and acting. To our knowledge, we are among the first to explore dense 3D representations as memory for embodied 3D LLMs — addressing a significant gap in current research as noted in recent 3D memory studies~\citep{yang20243dmem3dscenememory}.
Unlike standard approaches that rely solely on context windows~\citep{3dllm,huang2024embodied, zhu2024llava}, \model implements a dual-memory system: a limited-capacity working memory for current observations and an expandable episodic memory that stores past spatial-temporal information as dense 3D representations. The key innovation is our memory fusion module that actively integrates information from both memory systems based on task relevance and spatial-temporal relationships. 
This allows the model to leverage the benefits of dense 3D representations while mitigating their computational demands, maintaining coherent spatial-temporal understanding across extended task horizons. The fusion process
% intelligently 
preserves critical spatial relationships while accounting for their evolvement through agent interactions over time.

We evaluate popular 3D-LLMs and memory mechanisms on \benchmark. Experimental results demonstrate \model significantly outperforms all existing approaches in both in-domain and in-the-wild embodied tasks. Notably, while the performance of other methods drops sharply in the challenging in-the-wild setting, our method remains robust, achieving an average success rate of 32.1\%—demonstrating strong generalization capabilities. As task complexity increases from simple to hard, all existing approaches degrade significantly, achieving only $\sim$5\% success rate in hard in-the-wild tasks. In contrast, \model maintains a strong performance of 27.8\%, demonstrating its scalability and effectiveness in managing longer-term memory representations.

Our contributions can be summarized as below:
\vspace{-2mm}
\begin{itemize}
[align=right,itemindent=0em,labelsep=2pt,labelwidth=1em,leftmargin=*,itemsep=0em]
\item We propose a novel task that requires agents to execute action chains while maintaining and utilizing long-term spatial-temporal memory.

% \item We propose \benchmark, a comprehensive benchmark comprising over 26,000 trajectories and 1,860 fine-grained embodied tasks—ranging from simple to hard—along with question-answering target memory changes across time and space and captioning tasks, all designed to evaluate an agent's ability to reason over long-term memory in 3D environments.
\item We construct \benchmark, a comprehensive benchmark comprising over 26,000 trajectories and 1,860 fine-grained long-term memory embodied tasks—ranging from simple to hard—along with question-answering tasks that target memory changes across time and space, and captioning tasks in complex 3D environments.

\item We propose \model, an embodied 3D LLM with a novel memory fusion module for spatial-temporal reasoning, planning, and acting-which utilizes working memory tokens as queries to selectively fuse relevant features from episodic memory for efficient, task-aware decision-making.
\item Experimental results on embodied tasks, question-answering, and captioning demonstrate that  \model outperforms baselines by a large margin. 

\end{itemize}

\section{The Embodied 3D Long-Term Spatial-Temporal Memory Benchmark}
\vspace{-1mm}
\begin{figure}[t]
  \centering
  % \fbox{\rule[-.5cm]{0cm}{4cm} \rule[-.5cm]{4cm}{0cm}}
\includegraphics[trim=0cm 9.7cm 2.5cm 0cm, clip, width=\linewidth]{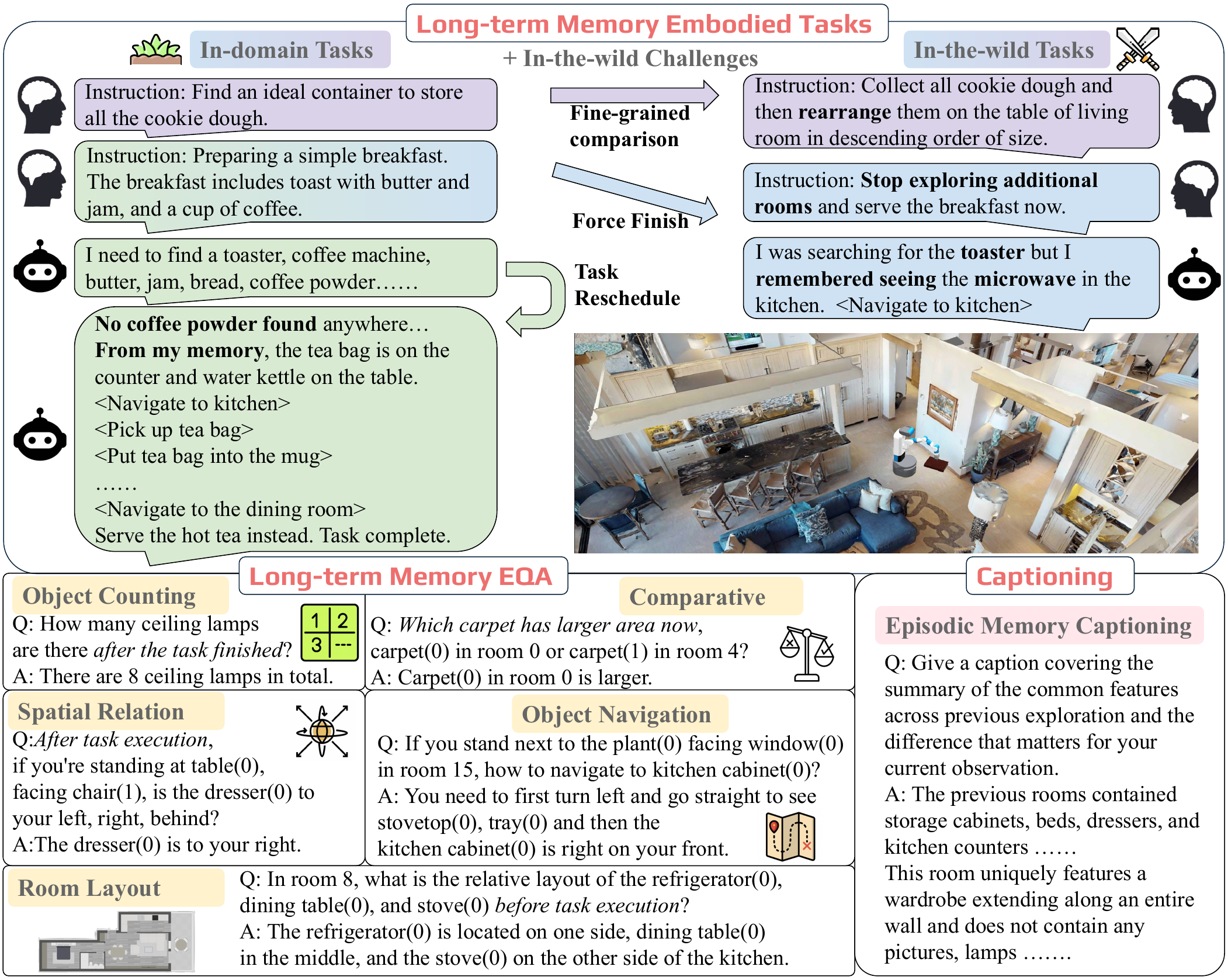}
  \caption{Overview of \benchmark. For long-term memory embodied tasks, we further incorporate in-the-wild challenges to test 3D agent's generalization abilities. Text inside < > indicates high-level action tokens. For complete embodied task trajectories, please refer to Appendix~\ref{appendix: data examples}.}
 \label{fig:benchmark}
 \vspace{-3mm}
\end{figure}

\subsection{Overview of \benchmark}
\label{sec:benchmark intro}
\begin{table*}[t]
\centering
% \small
\vspace{-10pt}
\vspace{1mm}
\renewcommand\tabcolsep{2pt}
\renewcommand\arraystretch{1}
% \resizebox{0.9\linewidth}{!}{
\resizebox{1\linewidth}{!}{
\begin{tabular}{@{}lcccccc@{}}
\toprule
 Benchmark  & \#Test Tasks & \#Train Trajectories  & Long-term Memory & Fine-grained complexity  & EQA & Captioning \\
\midrule

ALFWorld~\citep{ALFWorld20}    & 274 & 3,553 & \redx  & \redx    & NA  & NA\\
% VLMbench \cite{zheng2022vlmbench} & Manipulation   & Low     & 1 &    4760      & \greencheck & \redx    & \redx   \\
Behavior-1K~\citep{li2024behavior1khumancenteredembodiedai}    &1,000      & NA &  \redx  & \redx & NA  & NA \\
% GOAT-bench \cite{khanna2024goat} & Navigation &  Low & 1 & 3919 & \greencheck & \redx &  \redx \\
% AgentBench~\cite{liu2023agentbench}         & Multi-domain$^1$   & High    & 8 & 1091       & \redx     & \redx    & \greencheck     \\ 
VisualAgentBench~\citep{liu2024visualagentbench}   & 746  & 4,482 & \redx & \redx & NA  & NA \\
% Embodied Agent Interface~\cite{li2024embodied}         & Household   & High   & 2 & 438    & \redx   & \greencheck   & \greencheck    \\
% VLABench  \cite{zhang2024vlabench}  & Manipulation   & Low$^2$   & 1 & 100    & \greencheck   & \greencheck   & \greencheck    \\
EmbodiedBench~\citep{yang2025embodiedbenchcomprehensivebenchmarkingmultimodal}  & 1,128    & NA & \redx & \redx & NA  & NA \\

\midrule
\textbf{\benchmark(ours)    } &   1,860 & 26,276 & \greencheck & \greencheck & 865 & 167  \\
\bottomrule
\end{tabular}
}
\caption{Comparison with related benchmarks. \benchmark focus on spatial-temporal memory through fine-grained embodied tasks and EQA that span multiple ``pieces'' of long-term memory, distinguishing it from prior benchmarks that typically target single-step or short-horizon reasoning. Fine-grained complexity indicates our embodied task spans from simple to medium to hard.}
\label{table:comparison}
\vspace{-1mm}
\end{table*}

\paragraph{Design principles}
Long-term memory~\citep{camina2017neuroanatomical, friedman2018long, zlotnik2019memory} can be categorized into \textit{explicit memory} and \textit{implicit memory}. Explicit memory includes \textit{semantic memory}, which stores general knowledge and facts about the world, and \textit{episodic memory}, which consists of personal experiences that are time-stamped and context-specific. In contrast, implicit memory primarily involves \textit{procedural memory}, such as learned skills and habits.

To comprehensively evaluate 3D long-term memory for real-world applications, we design \benchmark following three core task categories: embodied tasks, long-term memory EQA, and captioning. As illustrated in Figure~\ref{fig:benchmark}, \textit{embodied tasks} require an embodied agent to solve realistic indoor environment challenges by leveraging both implicit and explicit long-term memory. \textit{Long-term memory EQA} tests the agent's ability to answer complex embodied questions using spatial-temporal memory. This task includes five subcategories: spatial reasoning questions, long-term object navigation, comparative reasoning, multi-room layout understanding, and semantic object counting. \textit{Captioning} tasks involve summarizing the agent's episodic memory to highlight shared and distinctive features across experiences, enabling more informed decision-making under the current task context.

% Statistics details in the appendix: QA 865 , caption: 167
% simple: domain: 177, wild: 177 x 3
% meidum: 173 ; 173 x 3
% hard: 112 ; 116 x3
% train: 13421, 7689, 5166 -> 26276

\subsection{Data Collection}
\label{sec:data collection}

\paragraph{Base environment construction} %Or Scene Generation? 

We build our scenes on top of the Habitat-Matterport 3D (HM3D) semantics dataset~\citep{ramakrishnan2021hm3d}, which has 1000 3D spaces and 10,600 rooms within those spaces. Pre-processing for the axis-aligned bounding box and using valid semantic label annotation, we filter to 182 3D spaces and 2,602 rooms. However, existing objects in HM3D scene are not interactive in Habitat-sim~\citep{szot2021habitat}. To expand our task diversity and enable embodied tasks, we add interactive objects from Objaverse~\citep{objaverse} which consists of 800K 3D objects spanning rich categories. More environment construction details are illustrated in Appendix~\ref{appendix: env construction}.

\paragraph{Generating task trajectories} 
Following~\citet{3dllm, Hong_2024_CVPR}, we adopt box-demonstration-instruction-based prompting, which utilizes the axis-aligned bounding boxes (AABB) of both rooms and objects within the 3D scenes to prompt Gemini~\citep{team2023gemini} to generate diverse tasks. We further prompt Gemini to incorporate interactive objects based on task requirements and their appropriateness within indoor environments. Detailed prompt instructions and few-shot demonstration examples are provided in Appendix~\ref{appendix: generation prompts}. To ensure the validity of the generated trajectories, we develop a trajectory simulation pipeline that verifies each trajectory step-by-step. At every step, the simulator checks: (1) the correctness of the agent’s location, (2) the existence and validity of referenced objects, and (3) the correctness of pick-up and put-down actions. Finally, we ensure that high-level actions can be executed in the simulator,  following~\citep{szot2023large, yang2025embodiedbenchcomprehensivebenchmarkingmultimodal}. Details of this implementation are in Appendix~\ref{appendix: traj validation}. On average, our filtering process yields a validation rate of approximately 24\%, ensuring the correctness and feasibility of the generated trajectories.

% Longterm embodied tasks necessarily involves of long chain of actions. We provide a list of legal actions in the environment to Gemini which generates a long pseudo ground-truth action trajectory. Inspired by slow thinking mechanism {cite cite} and to better help agent understand its performed actions, we also generate long chain-of-thought reasoning along with chain-of-actions. \wh{this may need one example to illustrates, refer it back to figure 1}. ... 
% \wh{I removed this, because we haven't draw or mention this in the figure.}

% If an error occurs, the trajectory is marked invalid and discarded. Only trajectories successfully completing all steps without errors are retained for subsequent training and evaluation. (yanjun)

\paragraph{Embodied data collection}
In our task settings, an embodied agent first performs random exploration within the environment to collect RGB-D observations and corresponding camera poses. Then the agent follows the task trajectory, incrementally exploring new environments, executing interaction actions, and receiving feedback with new RGB-D observation data. All interaction results are recorded and the reconstructed point cloud data is precomputed and stored locally to enable faster loading during both training and inference.
% Specifically, at each step of the data collection process, we sample a navigable point and make the agent move to the point along the shortest path. When the agent has arrived at a point, we rotate the agent 30◦ along z-axis for 12 times so that the agent can observe the 360◦ view of the scene at the position. It can also look up and down, with a random mild angle from [−10◦,10◦] along the x-axis. A picture is taken each time the agent rotates to a new orientation.

\subsection{Data Curation}
\label{sec: data curation}
\vspace{-1mm}

As mentioned previously, we collect embodied data by prompting Gemini. To enable a fine-grained analysis of long-term memory capacity, we divide the tasks into three subcategories: \textit{simple}, \textit{medium}, and \textit{hard}, comprising of 3, 5 and 10 multi-room scene settings respectively. 
%The \textit{simple} setting includes memory from 3 multi-room scenes, \textit{medium} includes 5, and \textit{hard} consists of 10 multi-room scenes. 
In total, we collect 51K trajectories, with 31K in the simple setting, 10K in the medium, and 10K in the hard.

To construct in-domain evaluation sets, we first remove training tasks and filter for instances that never shown in the agent’s working memory. For the in-the-wild evaluation set, we apply additional filtering to assess the agent’s generalization capabilities. Specifically, we select instances involving unseen objects and entirely unseen memory context, and we introduce novel in-the-wild challenges that differ from those encountered during training, as illustrated in Figure~\ref{fig:benchmark}.

For EQA data curation, we extract complete trajectories explored by agents and then prompt Gemini to generate question-answer pairs. %targeting key reasoning capabilities.
The questions are categorized into spatial reasoning, long-term object navigation, comparative reasoning, multi-room layout understanding, and semantic object counting. %, as described in $\S$\ref{sec:benchmark intro}. 
As shown in Figure~\ref{fig:benchmark}, these questions evaluate models on spatial-temporal changes in memory during embodied task execution. For long-term memory captioning, which primarily targets semantic episodic memory, we collect data across multiple rooms before and after the execution of each trajectory, enabling comparison and summarization of memory-relevant experiences.

\vspace{-2mm}

\paragraph{Quality control}
After constructing the entire benchmark, we implement two quality control procedures: automatic validation using trajectory simulation rules and a manual review of each benchmark instance. The automatic check involves re-running the trajectory simulation validation pipeline, as described in $\S$\ref{sec:data collection}, particularly for the in-the-wild tasks. For human validation, four student experts in the field manually inspect each benchmark example. We render multi-view images of the entire scene using the simulator and verify whether the benchmark annotations accurately correspond to the simulated environment.
More details are in Appendix~\ref{appendix: human validation}.

\vspace{-2mm}

% \section{\textsc{3DLLM-Mem}}
\section{3D Long-Term Spatial-Temporal Memory Model (\model)}
\vspace{-1mm}
\label{sec: method}

\begin{figure}[t]
  \centering
  % \fbox{\rule[-.5cm]{0cm}{4cm} \rule[-.5cm]{4cm}{0cm}}
% \includegraphics[trim=0cm 23.7cm 4.3cm 0cm, clip, width=\linewidth]{imgs/long_3dmem_method_figure.pptx-4.pdf}
\includegraphics[trim=0cm 23cm 4.3cm 0cm, clip, width=\linewidth]{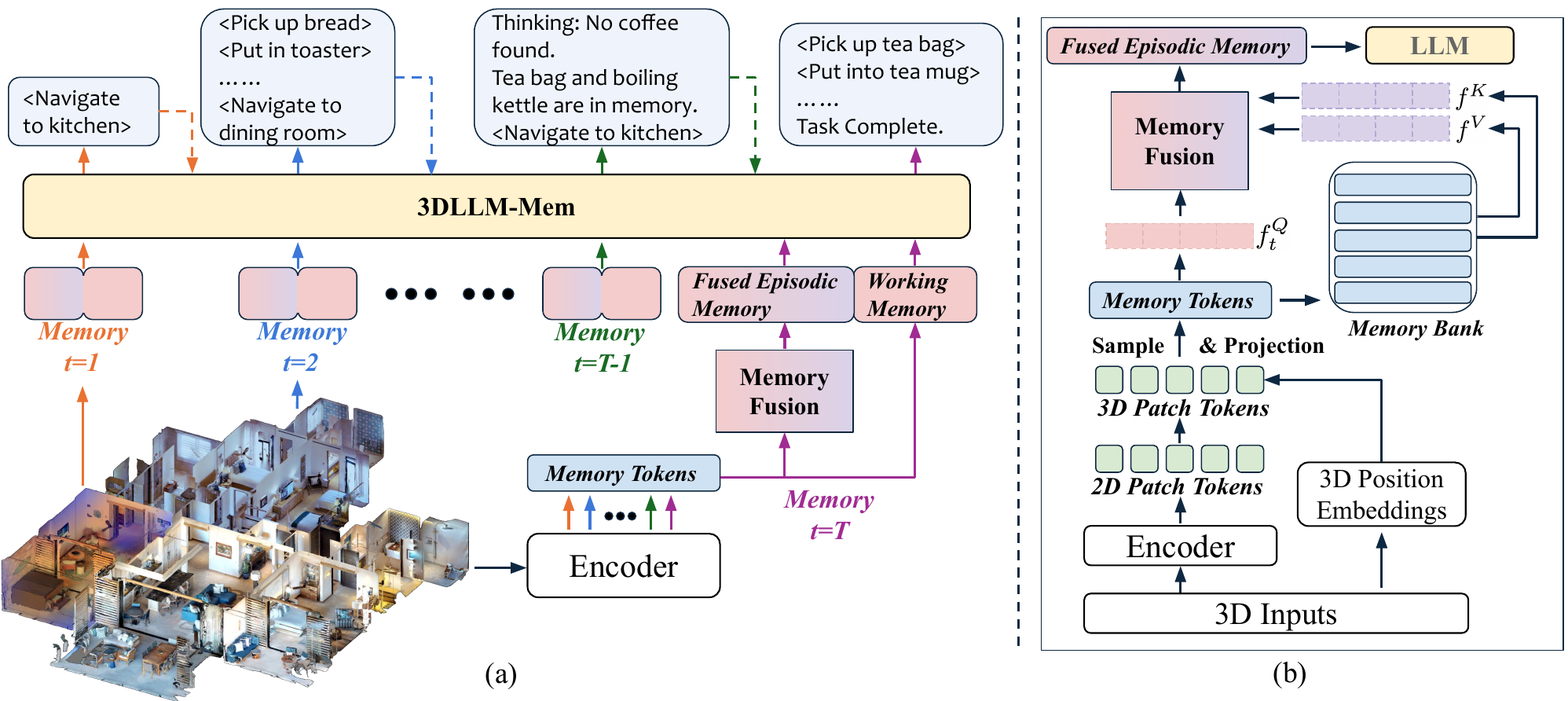}
  \caption{{(a)} We propose \model, a memory-enhanced 3D embodied agent that gradually form its long-term memory while executing tasks. Multiple timesteps are shown together but in different colors, with each timestep’s memory including the prior one. The task is ``prepare a simple breakfast'' as shown in Figure~\ref{fig:benchmark}. {(b)} Overview of our memory fusion mechanism. }
 \label{fig:method}
 \vspace{-4mm}
\end{figure}

\subsection{Preliminary} 
\vspace{-1mm}

Recent work on 3D Large Language Models (3D-LLMs) has showcased robust capabilities. We choose LLaVA-3D~\citep{zhu2024llava} as the base model to build our long-term memory 3D-LLM. LLaVA-3D directly builds on 2D-LLM with multi-view images as input and utilizing the 3D position embeddings to bring the 2D patches within a 3D spatial context to construct 3D patches. For each frame image, a CLIP encoder splits the image $X \in \mathbb{R}^{3 \times W \times H}$ into patches at the patch size $P$. For each 3D scene, $V$ multi-view image patch features are encoded and then projected into LLM space as $X_p \in \mathbb{R}^{V \times d \times w \times h}$, where $h = \left\lfloor \frac{H}{P} \right\rfloor, w = \left\lfloor \frac{W}{P} \right\rfloor$, and $d$ represents LLM's hidden dimension. 
The 3D positions in the 3D world are obtained with known depth image, camera intrinsic and extrinsic parameters and are further encoded into 3D position embeddings $P \in \mathbb{R}^{V \times d \times w \times h}$. These are directly added to the 2D patch visual tokens $X_p$, resulting in pixel-aligned 3D patches $X_{3D} \in \mathbb{R}^{V \times d \times w \times h }$. To reduce redundancy in 3D patches, we adopt the Farthest Point Sampling (FPS) strategy to downsample the 3D features to a fixed number of tokens, resulting in $X_{\text{3D Feat}} \in \mathbb{R}^{N \times d}$.

% \wh{let me know if this is lengthy and need to be shorten. }

% \subsection{Dynamic Memory Management}
\subsection{\model Memory Module}

% A 3D embodied agent gradually explores the environment, taking observations and interacting with the environment. For human, the current observations are stored in \textit{working memory}, as the exploration goes longer, the past observations are stored into the \textit{episodic memory}. Inspired by human mindset, Our 3D memory agent is also designed in this way. The current observation  $X^{[t=i]} \in \mathbb{R}^{N \times d}$ always in the context window, acting as the \textit{working memory}. As the agent accumulates more observations, the previous observations $X^{[t=1:T]} \in \mathbb{R}^{T \times N \times d}$ from timestep 1 to T are stored in \textit{episodic memory}. We propose to use a memory feature bank to save the episodic memory. For each observation at time step $j$, where $1 \leq j \leq T$, we first use a MLP layer to project the observation into memory feature space and then save it in the memory bank. 

A 3D embodied agent gradually explores the environment by collecting observations and interacting with surrounding environments. For humans, current observations are held in \textit{working memory}, while longer-term observations and experiences are stored in \textit{episodic memory}. Inspired by human cognitive structure, \model is designed with a similar paradigm as illustrated in Figure~\ref{fig:method}. The current observation at time step $t=i$, denoted as $X^{[t=i]} \in \mathbb{R}^{N \times d}$, remains within the context window and serves as the agent's \textit{working memory}. As the agent accumulates more experiences, past observations from time steps $1$ to $T$, represented as $X^{[t=1:T]} \in \mathbb{R}^{T \times N \times d}$, are stored as part of its \textit{episodic memory}, where $T$ denotes the total number of timesteps.

\paragraph{Episodic memory} To manage episodic memory, we propose the use of a memory feature bank. For each observation at time step $j$, where $1 \leq j \leq T$, we first apply a multi-layer perceptron (MLP) layer to project the observation into a memory-specific feature space, which is then stored in the memory bank for future retrieval. To further enhance the temporal understanding of the agent’s exploration, we incorporate sinusoidal positional embeddings to encode each time step $t=j$, and then directly added to the corresponding memory feature representations.

\paragraph{Memory fusion}

Our motivation is that an agent should leverage its current observations to recall the most relevant information from its episodic memory in order to complete the current task. To achieve this, we propose a mechanism called \textit{3D memory fusion}. Specifically, we encode the 3D features from the working memory into a shared memory space and use this representation as the query feature, denoted as $f_t^Q \in \mathbb{R}^{N \times M}$, where $M$ is the dimensionality of the memory feature space.

The episodic memory bank stores the corresponding key and value features from past observations: $f^K \in \mathbb{R}^{T \times N \times M}$ and $f^V \in \mathbb{R}^{T \times N \times M}$, respectively. Here, $T$ is the number of past timesteps and $N$ is the number of memory tokens per timestep. This structure allows the agent to retrieve task-relevant information through memory-query attention. The fused memory feature is then concatenated with the working memory feature to produce the final memory-enhanced representation  $f^M$ for the agent: 

% \begin{equation}
%     f_{fuse}^{Q} = \mathrm{Softmax}(\frac{f_{t}^Q (f^K)^\top}{\sqrt{C}})f^V, f^M = \mathrm{Concat}[f_{fuse}^{Q} ; f_{t}^{Q}]
%     \label{eq:attn_fuse}
% \end{equation}

% \begin{equation}
%     f_{\text{fuse}}^{Q} = \mathrm{Softmax}(\frac{f_{t}^Q (f^K)^\top}{\sqrt{C}}) f^V, \quad
%     f^{M} = \mathrm{Concat}[f_{\text{fuse}}^{Q}; f_{t}^{Q}]
%     \label{eq:attn_fuse}
% \end{equation}

\begin{equation}
    f_{\text{fuse}}^{Q} = \mathrm{Softmax}\left( \frac{f_{t}^Q (f^K)^\top}{\sqrt{C}} \right) f^V, \quad
    f^{M} = \mathrm{Concat}\left[ f_{\text{fuse}}^{Q}; f_{t}^{Q} \right]
    \label{eq:attn_fuse}
\end{equation}

\paragraph{Memory update} 
The working memory is dynamic and updated online. As the agent interacts with the environment, changes in the environment are immediately reflected in the working memory through updated 3D representations. When the agent moves to a new environment, the previous working memory is transferred to the episodic memory bank. If the corresponding environment already exists in the memory bank and has been modified by the agent, the memory entry is updated accordingly. Thus, the memory bank remains dynamic and reflects the latest state of the explored environments. As described in $\S$\ref{sec:data collection}, environment changes and corresponding observations are pre-collected and stored locally to facilitate efficient data loading during both training and inference.

% \subsubsection{Working Memory}
% \subsubsection{Episodic Memory}
% \subsubsection{Memory Fusion}

% \subsection{Embodied Action Chains}
% \wh{a small section or no}
% \subsubsection{Action Tokens}
% \subsubsection{Observation Tokens}

% \wh{our action tokens are pickup / put down and explore room and go to which room, do we need a small section for this?}

% \subsection{Training \& Inference}

\section{Experiments}
\vspace{-2mm}

In this section, we first introduce the experimental setup and existing memory management baselines in $\S$\ref{sec: experimental setup}. Then, we benchmark existing approaches on \benchmark, and present comprehensive results on embodied tasks, EQA, and captioning tasks to demonstrate the
effectiveness of our \model in $\S$\ref{sec: results}, along with qualitative results. Finally, in $\S$\ref{sec: ablate}, we conduct an ablation study of key design choices in \model, demonstrating the effectiveness of our proposed memory fusion mechanism. 
% Finally, in $\S$\ref{sec: qualitative}, we present qualitative results of \model to illustrate its effectiveness. 

\begin{table*}[t]
% \vspace{-3mm}
\begin{subtable}{1\textwidth}
\centering
 % \small
 \renewcommand\tabcolsep{2.5pt} % column space
 \renewcommand\arraystretch{0.95} % row space
 \resizebox{1.0\linewidth}{!}{
    % \begin{tabular}{l|llll|llll|llll|llll}
    \begin{tabular}{c|cccc|cccc|cccc|cccc}

    \toprule
    \multicolumn{1}{c|}{\multirow{4}{*}{Model}}   &\multicolumn{4}{c|}{Simple} & \multicolumn{4}{c|}{Medium} & \multicolumn{4}{c|}{{Hard}}  & \multicolumn{4}{c}{{Average}}  \\
      \cmidrule(lr){2-5}  \cmidrule(lr){6-9} \cmidrule(lr){10-13} \cmidrule(lr){14-17}
     &\multicolumn{2}{c}{In-domain}  & \multicolumn{2}{c}{In-the-wild}  &\multicolumn{2}{c}{In-domain}  & \multicolumn{2}{c}{In-the-wild}&\multicolumn{2}{c}{In-domain}  & \multicolumn{2}{c}{In-the-wild}&\multicolumn{2}{c}{In-domain}  & \multicolumn{2}{c}{In-the-wild}\\ 
     \cmidrule(lr){2-3}  \cmidrule(lr){4-5} \cmidrule(lr){6-7} \cmidrule(lr){8-9}\cmidrule(lr){10-11}\cmidrule(lr){12-13}\cmidrule(lr){14-15}\cmidrule(lr){16-17}
      & \header{SR} & \header{Sub-SR}  & \header{SR} & \header{Sub-SR}  & \header{SR} & \header{Sub-SR}  & \header{SR} & \header{Sub-SR}  & \header{SR} & \header{Sub-SR}  & \header{SR} & \header{Sub-SR}  & \header{SR} & \header{Sub-SR}  & \header{SR} & \header{Sub-SR} \\
    \midrule 
    % \rowcolor[rgb]{0.93,0.93,0.93} \multicolumn{11}{l}{\textit{Heuristic baselines}} \\
    3D-LLM (Finetuned)& 10.4 &20.3 & 9.1 &18.5 & - & - & - & - & - & - & - & - & - & - & - & -\\
    % 3D-Mem (GPT4-o) & \\
    % 3D-Mem (Gemma 3) & \\
    Everything in Context  & 35.5 &63.9 & 32.4& 45.2 & - & - & - & - & - & - & - & - & - & - & - & -\\
    Most Recent Memory &32.8 & 62.3 & 23.4 & 38.6 & 20.1 & 34.8 & 12.4 & 25.3 & 10.4 & 20.7 & 5.4 & 12.1 &21.1     &39.3 & 13.7 & 25.3 \\
    Retrieval-Augmented Memory & 34.2 & 63.0 &28.3 & 46.2 & 21.8 & 40.2 &13.7 & 28.0 &10.8 & 21.6  & 4.8 & 10.6 &22.3 & 41.6  &15.6  & 28.3 \\ 
    \midrule
     \rowcolor{black!10} 
    \model (Ours) &45.5 & 73.4  & 37.0 &65.4 & 36.8 & 67.8 &31.6 &57.4 &30.5 & 46.2 & 27.8 &42.1 &37.6       &62.5& 32.1 & 55.0\\
     
    \bottomrule
    \end{tabular}
    }
    % \vspace{-2mm}
    \caption{Results on \benchmark embodied tasks. SR stands for success rate. Sub-SR stands for sub-success rate. Our model outperforms existing approaches by a large margin.}
% \vspace{-3mm}
\label{tab:embodiedresults}
\end{subtable}

\begin{subtable}{1\textwidth}
% \sisetup{table-format=4.0} %

\centering
 % \small
 \renewcommand\tabcolsep{2.5pt} % column space
 \renewcommand\arraystretch{0.95} % row space
 \resizebox{1.0\linewidth}{!}{
    % \begin{tabular}{l|ll|lllll|lll|l}
    \begin{tabular}{c|cc|ccccc|ccc}

    \toprule
    \multicolumn{1}{c|}{\multirow{3}{*}{Model}}   &\multicolumn{2}{c|}{Embodied Task} & \multicolumn{5}{c|}{Embodied Question Answering (EQA)} & \multicolumn{3}{c}{{Captioning}}  \\
    % & \multicolumn{1}{c}{\multirow{2}{*}{Average}} \\
      \cmidrule(lr){2-3}  \cmidrule(lr){4-8} \cmidrule(lr){9-11}
     & \header{In-domain} & \header{In-the-wild} & \header{Spatial} & \header{Nav.} & \header{Comparative} & \header{Layout} & \header{Count} & \header{BLEU1} & \header{BLEU4} & \header{METEOR}   \\ 
    \midrule
    % \rowcolor[rgb]{0.93,0.93,0.93} \multicolumn{11}{l}{\textit{Heuristic baselines}} \\
    3D-LLM (Finetuned) & - & - & 2.9 &5.8& 0.0& 7.7&0.0  & 42.3 & 12.0 & 30.6\\
    3D-Mem (GPT4-o) & - & - & 39.9 &11.0&25.8& 19.1&7.8& 41.7 & 4.7 & 31.8 \\
    3D-Mem (Gemini-2.5-Flash) & - & - & 41.6 &18.2 & 37.6& 30.2&12.7 & 42.8 & 4.8 & 29.6 \\
    3D-Mem (Gemini-2.5-Pro) & - & - & 39.7 &27.7 &36.0 & 35.2& 16.4&41.5 & 3.0 & 28.6  \\
    Most Recent Memory & 21.1 & 13.7& 27.5&30.2 & 24.3& 20.1& 10.5& 32.4 & 10.1 & 25.6 \\
    Retrieval-Augmented Memory & 22.3 &15.6 &38.0 &33.4 &31.8 &29.7 &15.6 & 40.8 & 11.5 & 29.3 \\ 
    \midrule
     \rowcolor{black!10} 
  \model (Ours) &37.6  &32.1 &62.8 & 40.6&41.4 &39.9 &26.3 & 58.2 & 18.8 & 37.3 \\
     
    \bottomrule
    \end{tabular}
    }
    % \vspace{-2mm}
    \caption{Results on all tasks in \benchmark. Average success rate is reported for embodied tasks. \textit{Nav.} stands for long-term object navigation. We report accuracy score for open-ended EQA evaluation and follow the standard LLM-as-judge evaluation protocol by prompting Gemini. 
    Evaluation details are provided in Appendix~\ref{appendix: generation prompts}.
 }
    % you can use \high{20.5}  and  \best{50.5}  
% \vspace{-3mm}
\label{tab:mainresults}
\end{subtable}

\caption{Comparison with 3D memory models and standard memory management approaches. Our model, \model, achieves the best performance across embodied, EQA and captioning tasks.}
\vspace{-3mm}
\label{tab:openset}
\end{table*}
% \vspace{-2mm}

\subsection{Experimental Setup}
\label{sec: experimental setup}

\paragraph{Implementation details} 
We implement our model based on LLaVA-3D~\citep{zhu2024llava}, modifying it to be compatible with Google TPUs with PyTorch/XLA frameworks~\citep{pytorch, xla} . We first expand the model’s context window to 8192 tokens to accommodate long-term memory inputs. We then fine-tune our proposed memory module along with the LLM decoder using our training split, initializing from LLaVA-3D’s pretrained weights. Training is conducted on 8 Google Cloud TPU v5p cores with a batch size of 256. Our model is trained using supervised fine-tuning (SFT) with a standard language modeling loss. More details are provided in Appendix~\ref{appendix: implement details}.
%more detials in appendix. 

\paragraph{Baselines}
We compare \model against a broad range of memory management approaches:
\vspace{-2mm}
\begin{itemize}
[align=right,itemindent=0em,labelsep=2pt,labelwidth=1em,leftmargin=*,itemsep=0em]

\item \textbf{Everything in Context.} For a small subset of scenes, it is feasible to fit all observations directly into the model's context window.

\item \textbf{Most Recent Memory.} 
Since retaining all observations in context is infeasible, we keep only the most recent observations, assuming they are most relevant to the current task.

\item \textbf{Retrieval-Augmented Memory.} Inspired by retrieval-based techniques, we adopt a memory bank that stores past observations. During inference, the most relevant memory entries are retrieved and appended before the working memory to augment reasoning.

\item \textbf{3D-LLM}~\citep{3dllm}. A popular 3D LLM recognized by the community. We finetune it on our training data and report its performance using the ``everything in context'' strategy with the longest context window supported. Further details are provided in Appendix~\ref{appendix: eval setup}.

\item \textbf{3D-Mem}~\citep{yang20243dmem3dscenememory}. A framework designed for 3D scene memory in embodied exploration and reasoning. 
% It leverages informative multi-view images and frontier snapshots to guide exploration. 
However, this method does not support embodied interaction or action execution.
\end{itemize}

\subsection{Experimental Results}
\label{sec: results}

% \paragraph{Results on Embodied Tasks}
% As shown in Table~\ref{tab:embodiedresults}, 3DLLM-Mem is significantly better than all existing approaches in both in-domain and in-the-wild embodied tasks. More importantly, while other approaches performance degrades drastically for in-the-wild embodied tasks, our approach is still robust with a 32.1 total success rate on average for in-the-wild tasks, emphaissing the capabilities of generalization abilities. Both most recent memory approach and Retrieval-Augmented (RAG) memory approach perform poorly, with RAG memory only being slightly better which indicate the chanllegns of retrieving the correct episodic memory. Interestingly, everything in context demonstrate better performance than most recent and RAG memory, indicates the model can utilize all information well if everything can be fit in there context which is also intuitive. However, 3DLLM-Mem still signifcantly better than everything in context in simple scenarios, indicating the benefits of fusion the important and useful memory so that the model can better utilize it. From simple to hard task scenarios, exisiting appraoches all decrease drastically as the memory grows longer with only ~5\% on hard in-the-wild tasks comparing to our appaorch is 27.8, indicating our approach is suitable and scablabel for longer long-term memory mamnagement and reprentation.  

\paragraph{Results on embodied tasks}
As shown in Table~\ref{tab:embodiedresults}, \model significantly outperforms all existing approaches on both in-domain and in-the-wild embodied tasks. Notably, while the performance of other methods drops sharply in the in-the-wild setting, our method demonstrates strong generalization capabilities with a average success rate of 32.1\%. 3D-LLM showcases the lowest performance even under simple task settings, highlighting the necessity of incorporating an explicit memory module.
Both the \textit{Most Recent Memory} and \textit{Retrieval-Augmented Memory} (RAG) baselines perform poorly in this setting, with RAG showing only a slight improvement, highlighting the challenges of retrieving relevant episodic memory. Interestingly, the \textit{Everything in Context} baseline performs better than both recent memory and RAG approaches, suggesting that when all information can fit within the context window, the model can effectively utilize it. However, \model still outperforms \textit{Everything in Context}, indicating the benefits of selectively fusing task-relevant memory features to better guide embodied reasoning and execution. As task complexity increases from simple to hard, all existing approaches degrade significantly, achieving only $\sim$5\% success rate in hard in-the-wild tasks. In contrast, \model maintains a strong performance of 27.8\%, demonstrating its scalability and effectiveness in managing longer-term memory representations.
\vspace{-1mm}

\paragraph{Results on long-term EQA and captioning}
As shown in Table~\ref{tab:mainresults}, \model consistently outperforms all existing approaches across all tasks in our benchmark. Notably, \textsc{3D-LLM} achieves the second-best performance on the captioning task, highlighting its strong ability to summarize object-centric semantic memory. However, due to limited context length, it performs poorly on the EQA tasks, which require long-term spatial-temporal reasoning. 
% thus demonstrating the importance of a dedicated memory module.
3D-Mem demonstrates improved performance in EQA over other baseline approaches. However, it falls short on spatial relation, navigation and object counting tasks, indicating the limitation of relying solely on aggregated image-centric memories.
% Compared to \textit{Most Recent Memory} and \textit{RAG Memory} baselines, which show promising results on object counting and navigation tasks, \model significantly outperforms both, which further demonstrate the effectiveness of our memory fusion technique. 
\model significantly outperforms both \textit{Most Recent Memory} and \textit{RAG Memory}, which further demonstrates the effectiveness of our memory fusion technique. 

\vspace{-1mm}

\begin{figure}[t]
  \centering
\includegraphics[trim=0cm 30.2cm 10.8cm 0cm, clip, width=\linewidth]{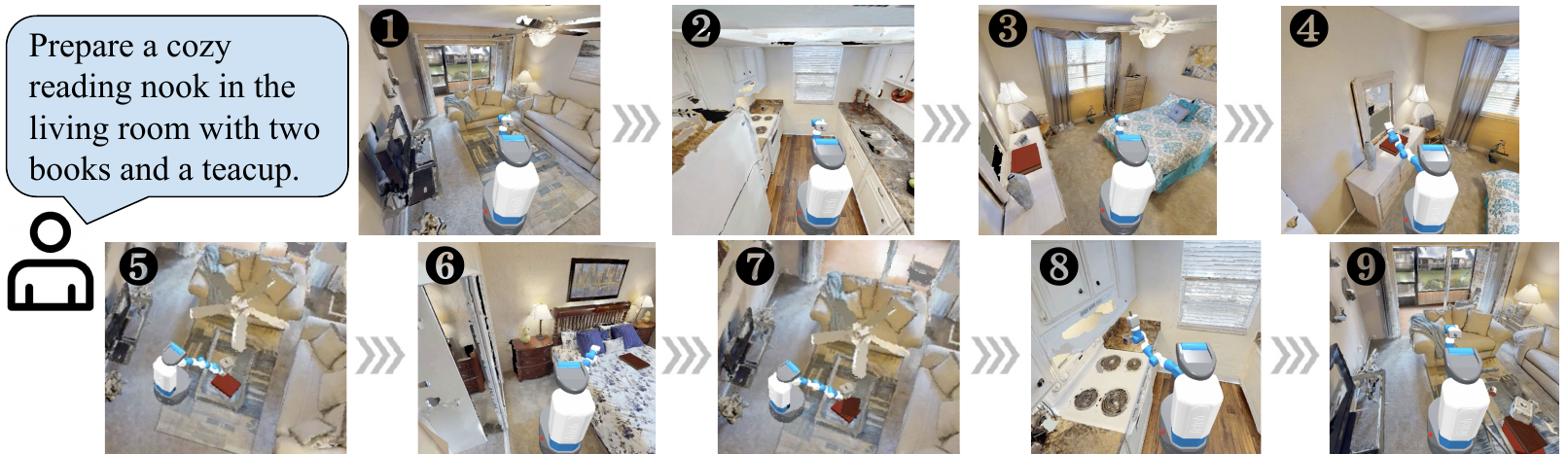}
  % \caption{Qualitative example of \model. The task instruction is: \textit{Prepare a cozy 
  \caption{
Qualitative example of \model, which maintains and utilizes a long-term memory to complete the task. Detailed task execution trajectory can be found in Figure~\ref{fig:qualitative}.
}
 \label{fig:qualitative paper}
 \vspace{-3mm}
\end{figure}

\paragraph{Qualitative results}
We provide qualitative examples in Figure~\ref{fig:qualitative paper} and a more detailed version with explanations in Figure~\ref{fig:qualitative} (Appendix~\ref{appendix: qualitative}), demonstrating that \model is capable of maintaining long-term memory and executing complex tasks in embodied environments. 
% Additional examples are included in the \textbf{supplementary materials}.

\vspace{-2mm}

\subsection{Ablation Study}
\label{sec: ablate}

Our approach initializes the fused memory using working memory features, aiming to fuse the most relevant memories for the current task. We ablate several design choices for initializing the fusion query, as shown in Table~\ref{tab:ablation}. When using either the most recent episodic memory or learnable zero parameters, performance degrades compared to our proposed method.
Interestingly, using the most recent memory outperforms zero initialization in the simple setting but underperforms in the hard setting. One possible explanation is that recent memory initialization encourages fusion with nearby observations, which may be sufficient for simple tasks and leads to faster convergence. In contrast, zero initialization is guided solely by training supervision to learn which memories are most useful.
In summary, the ablation results demonstrate that initializing fusion queries with working memory tokens provides the most effective and robust design choice for long-term memory fusion.

\vspace{-2mm}

\section{Related Works}
\vspace{-1mm}

\paragraph{3D Large Language Models}

3D Large Language Models (3D-LLMs) have demonstrated promising results across a wide variety of tasks, including 3D scene understanding, object detection, and segmentation~\citep{3dllm,zhou2023uni3d, huang2024chat, chen2024grounded3dllm, pointllm}. In parallel, 3D embodied agents have expanded these capabilities to planning and action in interactive environments~\citep{rt22023arxiv,huang2024embodied, Chen_2024_CVPR, black2024pi0visionlanguageactionflowmodel}. 
Yet, existing models face significant challenges when performing long-horizon embodied tasks in densely populated 3D environments that require reasoning over long-term spatial-temporal memory. To address this, we propose an explicit memory module inspired by the structure of human implicit and explicit memory. Our model employs a memory fusion mechanism that efficiently retrieves and learns task-relevant information, resulting in enhanced performance on complex embodied tasks.

% \subsection{3D-LLMs}
% \subsection{Long-term Embodied Planning}
% \subsection{Memory Retrieval}

\paragraph{Long-term Embodied Trajectories}

Embodied AI simulators \citep{Matterport3D, ai2thor, szot2021habitat, shen2021igibson} have fostered the development of embodied AI agents. Grounded in these environments, some existing benchmarks focus on high-level planning tasks, typically involving short trajectories that can often be completed within single-room settings, thereby requiring minimal spatial-temporal memory~\citep{Shridhar_2020_CVPR_alfred, ALFWorld20, li2024behavior1khumancenteredembodiedai, szot2023large,li2024embodied,yang2025embodiedbenchcomprehensivebenchmarkingmultimodal}. 
Other benchmarks emphasize long-term scene exploration with extended trajectories, but are primarily centered around navigation tasks and often lack embodied interaction support~\citep{RoboTHOR, ramakrishnan2021hm3d, krantz2022instancespecificimagegoalnavigation, khanna2024goatbench}.
To bridge this gap, we introduce \benchmark, a benchmark specifically designed to evaluate long-horizon task execution that requires rich spatial-temporal memory and full embodied task support, as summarized in Table~\ref{table:comparison}.

\paragraph{Embodied Question Answering Benchmark}

Embodied Question Answering (EQA) benchmarks~\citep{Das_2018_CVPR, wijmans2019embodiedquestionansweringphotorealistic, yu2019multitargetembodiedquestionanswering} have been developed to advance goal-driven agents that can perceive their environment. % and execute actions accordingly.
Some EQA benchmarks also include embodied memory QA evaluation, such as OpenEQA~\citep{OpenEQA2023}, which includes an episodic memory QA split, and~\citet{yang2024think}, which focuses on spatial memory QA. In contrast, our benchmark, \benchmark jointly targets both spatial and episodic memory, especially their changes over time, while also supporting embodied action tasks, EQA and captioning.
For specific comparison on EQA, our long-term memory EQA tasks are designed to require reasoning over multiple ``pieces'' of memory and their changes across time and space. 
Additionally, we consider the agent’s location in the scene at the moment of answering each question during evaluation.

\paragraph{Memory System} 

Memory is a fundamental component of AI systems, with early work in the context of LLM agents that utilize memory for decision-making in web-based and sandbox environments~\citep{shinn2023reflexion,DanyangZhang2023_Rememberer,packer2024memgptllmsoperatingsystems,zhang2024surveymemorymechanismlarge}. Most existing approaches construct an experience pool or memory bank and focus on improving the retrieval of useful past information~\citep{zhao2024expel, gao2024selfevolvinggptlifelongautonomous, xu2025mem}.
In the computer vision domain, temporal memory has been studied extensively in video understanding and generation tasks~\citep{wang2021temporal, diao2025temporalworkingmemoryqueryguided}, while spatial memory has been applied to scene-level visual understanding and 3D reconstruction~\citep{wang20243d,m3}.  Recent work such as 3D-Mem~\citep{yang20243dmem3dscenememory} has investigated 3D scene memory for exploration and reasoning by prompting vision-language models. In contrast, our work focuses on dense 3D memory representations that are critical for real-world embodied scenarios, where task execution depends heavily on maintaining and reasoning over long-term spatial-temporal memory.

% \wh{I put this paragraph because I want to evaluate our method in OpenEQA, then I find out I couldn't.}

\begin{table*}[t]
% \vspace{-3mm}
\centering
 % \small
 \renewcommand\tabcolsep{2.5pt} % column space
 \renewcommand\arraystretch{0.95} % row space
 \resizebox{1.0\linewidth}{!}{
    % \begin{tabular}{l|llll|llll|llll|llll}
    \begin{tabular}{c|cccc|cccc|cccc|cccc}

    \toprule
    \multicolumn{1}{c|}{\multirow{4}{*}{Model}}   &\multicolumn{4}{c|}{Simple} & \multicolumn{4}{c|}{Medium} & \multicolumn{4}{c|}{{Hard}}  & \multicolumn{4}{c}{{Average}}  \\
      \cmidrule(lr){2-5}  \cmidrule(lr){6-9} \cmidrule(lr){10-13} \cmidrule(lr){14-17}
     &\multicolumn{2}{c}{In-domain}  & \multicolumn{2}{c}{In-the-wild}  &\multicolumn{2}{c}{In-domain}  & \multicolumn{2}{c}{In-the-wild}&\multicolumn{2}{c}{In-domain}  & \multicolumn{2}{c}{In-the-wild}&\multicolumn{2}{c}{In-domain}  & \multicolumn{2}{c}{In-the-wild}\\ 
     \cmidrule(lr){2-3}  \cmidrule(lr){4-5} \cmidrule(lr){6-7} \cmidrule(lr){8-9}\cmidrule(lr){10-11}\cmidrule(lr){12-13}\cmidrule(lr){14-15}\cmidrule(lr){16-17}
      & \header{SR} & \header{Sub-SR}  & \header{SR} & \header{Sub-SR}  & \header{SR} & \header{Sub-SR}  & \header{SR} & \header{Sub-SR}  & \header{SR} & \header{Sub-SR}  & \header{SR} & \header{Sub-SR}  & \header{SR} & \header{Sub-SR}  & \header{SR} & \header{Sub-SR} \\
    \midrule 
    % \rowcolor[rgb]{0.93,0.93,0.93} \multicolumn{11}{l}{\textit{Heuristic baselines}} \\
    \model &45.5 & 73.4  & 37.0 &65.4 & 36.8 & 67.8 &31.6 &57.4 &30.5 & 46.2 & 27.8 &42.1 &37.6  &62.5& 32.1 & 55.0   \\
    \midrule 
    % Voxel-based feature  \\
    Init with Most Recent Episodic Memory& 42.3 & 69.4  & 28.6  & 50.7 & 32.4 & 58.6& 23.7 & 45.1 & 22.6 & 37.8 & 15.3 & 31.4 & 32.4 & 55.3 & 22.5 & 42.4  \\
    Init with Learnable Zero Parameters & 41.4 & 67.2 & 27.9 & 50.0 & 33.0 & 59.2 & 23.4 & 45.8 & 24.2 & 40.4 & 18.6 & 35.6 & 32.9 & 55.6 & 23.3 & 43.8    \\
    \bottomrule
    \end{tabular}
    }
    % \vspace{-2mm}
    \caption{Ablation study of query initialization designs in our memory fusion module.}
    % you can use \high{20.5}  and  \best{50.5}  
\vspace{-4mm}
\label{tab:ablation}
\end{table*}

\section{Conclusion}
\vspace{-2mm}
\label{sec: conclusion}

In this work, we introduce \benchmark, a comprehensive benchmark containing fine-grained long-term memory embodied tasks—ranging from simple to hard—along with question-answering tasks that target memory changes across time and space, and captioning task in complex 3D environments. 
We propose \model, an embodied 3D-LLM with novel memory fusion approach for spatial-temporal reasoning, planning, and acting. One limitation of our model is that currently \model does not involve low-level navigation and control policy, but utilizes high-level pre-defined policies in simulator for carrying out the actions. We think that such aspects are orthogonal to our study, and could be explored and seamlessly integrated into our framework in the future.

\begin{ack} %%%%%%%%comment: this is neurips style 
We thank anonymous reviewers and other members
of UCLA-NLP+ group for their helpful comments.
This work was partially supported by U.S. DARPA
ECOLE Program No. \#HR00112390060, ONR grant
N00014-23-1-2780, Amazon Research Award, and a Google gift fund. Peng and Chang have financial COI with Google and Amazon and were supported in part by a grant from DARPA to the Simons Institute for the Theory of Computing.
\end{ack}

\bibliography{custom}

\begin{thebibliography}{59}
\expandafter\ifx\csname natexlab\endcsname\relax\def\natexlab#1{#1}\fi

\bibitem[{Black et~al.(2024)Black, Brown, Driess, Esmail, Equi, Finn, Fusai, Groom, Hausman, Ichter, Jakubczak, Jones, Ke, Levine, Li-Bell, Mothukuri, Nair, Pertsch, Shi, Tanner, Vuong, Walling, Wang, and Zhilinsky}]{black2024pi0visionlanguageactionflowmodel}
Kevin Black, Noah Brown, Danny Driess, Adnan Esmail, Michael Equi, Chelsea Finn, Niccolo Fusai, Lachy Groom, Karol Hausman, Brian Ichter, Szymon Jakubczak, Tim Jones, Liyiming Ke, Sergey Levine, Adrian Li-Bell, Mohith Mothukuri, Suraj Nair, Karl Pertsch, Lucy~Xiaoyang Shi, James Tanner, Quan Vuong, Anna Walling, Haohuan Wang, and Ury Zhilinsky. 2024.
\newblock \href {https://arxiv.org/abs/2410.24164} {$\pi_0$: A vision-language-action flow model for general robot control}.

\bibitem[{Brohan et~al.(2023)Brohan, Brown, Carbajal, Chebotar, Chen, Choromanski, Ding, Driess, Dubey, Finn, Florence, Fu, Arenas, Gopalakrishnan, Han, Hausman, Herzog, Hsu, Ichter, Irpan, Joshi, Julian, Kalashnikov, Kuang, Leal, Lee, Lee, Levine, Lu, Michalewski, Mordatch, Pertsch, Rao, Reymann, Ryoo, Salazar, Sanketi, Sermanet, Singh, Singh, Soricut, Tran, Vanhoucke, Vuong, Wahid, Welker, Wohlhart, Wu, Xia, Xiao, Xu, Xu, Yu, and Zitkovich}]{rt22023arxiv}
Anthony Brohan, Noah Brown, Justice Carbajal, Yevgen Chebotar, Xi~Chen, Krzysztof Choromanski, Tianli Ding, Danny Driess, Avinava Dubey, Chelsea Finn, Pete Florence, Chuyuan Fu, Montse~Gonzalez Arenas, Keerthana Gopalakrishnan, Kehang Han, Karol Hausman, Alex Herzog, Jasmine Hsu, Brian Ichter, Alex Irpan, Nikhil Joshi, Ryan Julian, Dmitry Kalashnikov, Yuheng Kuang, Isabel Leal, Lisa Lee, Tsang-Wei~Edward Lee, Sergey Levine, Yao Lu, Henryk Michalewski, Igor Mordatch, Karl Pertsch, Kanishka Rao, Krista Reymann, Michael Ryoo, Grecia Salazar, Pannag Sanketi, Pierre Sermanet, Jaspiar Singh, Anikait Singh, Radu Soricut, Huong Tran, Vincent Vanhoucke, Quan Vuong, Ayzaan Wahid, Stefan Welker, Paul Wohlhart, Jialin Wu, Fei Xia, Ted Xiao, Peng Xu, Sichun Xu, Tianhe Yu, and Brianna Zitkovich. 2023.
\newblock \href {https://arxiv.org/abs/2307.15818} {Rt-2: Vision-language-action models transfer web knowledge to robotic control}.
\newblock volume abs/2307.15818.

\bibitem[{Camina and Güell(2017)}]{camina2017neuroanatomical}
Eduardo Camina and Francisco Güell. 2017.
\newblock \href {https://doi.org/10.3389/fphar.2017.00438} {The neuroanatomical, neurophysiological and psychological basis of memory: Current models and their origins}.
\newblock \emph{Frontiers in Pharmacology}, 8:438.

\bibitem[{Chang et~al.(2017)Chang, Dai, Funkhouser, Halber, Niessner, Savva, Song, Zeng, and Zhang}]{Matterport3D}
Angel Chang, Angela Dai, Thomas Funkhouser, Maciej Halber, Matthias Niessner, Manolis Savva, Shuran Song, Andy Zeng, and Yinda Zhang. 2017.
\newblock Matterport3d: Learning from rgb-d data in indoor environments.
\newblock \emph{International Conference on 3D Vision (3DV)}.

\bibitem[{Chen et~al.(2024{\natexlab{a}})Chen, Xu, Kirmani, Ichter, Sadigh, Guibas, and Xia}]{Chen_2024_CVPR}
Boyuan Chen, Zhuo Xu, Sean Kirmani, Brian Ichter, Dorsa Sadigh, Leonidas~J. Guibas, and Fei Xia. 2024{\natexlab{a}}.
\newblock \href {https://doi.org/10.1109/CVPR52733.2024.01370} {Spatialvlm: Endowing vision-language models with spatial reasoning capabilities}.
\newblock In \emph{{IEEE/CVF} Conference on Computer Vision and Pattern Recognition, {CVPR} 2024, Seattle, WA, USA, June 16-22, 2024}, pages 14455--14465. {IEEE}.

\bibitem[{Chen et~al.(2024{\natexlab{b}})Chen, Yang, Huang, Wang, Lyu, Xu, Lin, and Pang}]{chen2024grounded3dllm}
Yilun Chen, Shuai Yang, Haifeng Huang, Tai Wang, Ruiyuan Lyu, Runsen Xu, Dahua Lin, and Jiangmiao Pang. 2024{\natexlab{b}}.
\newblock \href {https://arxiv.org/abs/2405.10370} {Grounded 3d-llm with referent tokens}.
\newblock \emph{ArXiv preprint}, abs/2405.10370.

\bibitem[{Chung et~al.(2022)Chung, Hou, Longpre, Zoph, Tay, Fedus, Li, Wang, Dehghani, Brahma et~al.}]{chung2022scaling}
Hyung~Won Chung, Le~Hou, Shayne Longpre, Barret Zoph, Yi~Tay, William Fedus, Yunxuan Li, Xuezhi Wang, Mostafa Dehghani, Siddhartha Brahma, et~al. 2022.
\newblock \href {https://arxiv.org/abs/2210.11416} {Scaling instruction-finetuned language models}.
\newblock \emph{ArXiv preprint}, abs/2210.11416.

\bibitem[{Das et~al.(2018)Das, Datta, Gkioxari, Lee, Parikh, and Batra}]{Das_2018_CVPR}
Abhishek Das, Samyak Datta, Georgia Gkioxari, Stefan Lee, Devi Parikh, and Dhruv Batra. 2018.
\newblock \href {https://doi.org/10.1109/CVPR.2018.00008} {Embodied question answering}.
\newblock In \emph{2018 {IEEE} Conference on Computer Vision and Pattern Recognition, {CVPR} 2018, Salt Lake City, UT, USA, June 18-22, 2018}, pages 1--10. {IEEE} Computer Society.

\bibitem[{Deitke et~al.(2020)Deitke, Han, Herrasti, Kembhavi, Kolve, Mottaghi, Salvador, Schwenk, VanderBilt, Wallingford, Weihs, Yatskar, and Farhadi}]{RoboTHOR}
Matt Deitke, Winson Han, Alvaro Herrasti, Aniruddha Kembhavi, Eric Kolve, Roozbeh Mottaghi, Jordi Salvador, Dustin Schwenk, Eli VanderBilt, Matthew Wallingford, Luca Weihs, Mark Yatskar, and Ali Farhadi. 2020.
\newblock \href {https://doi.org/10.1109/CVPR42600.2020.00323} {Robothor: An open simulation-to-real embodied {AI} platform}.
\newblock In \emph{2020 {IEEE/CVF} Conference on Computer Vision and Pattern Recognition, {CVPR} 2020, Seattle, WA, USA, June 13-19, 2020}, pages 3161--3171. {IEEE}.

\bibitem[{Deitke et~al.(2023)Deitke, Schwenk, Salvador, Weihs, Michel, VanderBilt, Schmidt, Ehsani, Kembhavi, and Farhadi}]{objaverse}
Matt Deitke, Dustin Schwenk, Jordi Salvador, Luca Weihs, Oscar Michel, Eli VanderBilt, Ludwig Schmidt, Kiana Ehsani, Aniruddha Kembhavi, and Ali Farhadi. 2023.
\newblock \href {https://doi.org/10.1109/CVPR52729.2023.01263} {Objaverse: {A} universe of annotated 3d objects}.
\newblock In \emph{{IEEE/CVF} Conference on Computer Vision and Pattern Recognition, {CVPR} 2023, Vancouver, BC, Canada, June 17-24, 2023}, pages 13142--13153. {IEEE}.

\bibitem[{Diao et~al.(2025)Diao, Zhang, Wu, Ouyang, Qing, Cheng, Vosoughi, and Gui}]{diao2025temporalworkingmemoryqueryguided}
Xingjian Diao, Chunhui Zhang, Weiyi Wu, Zhongyu Ouyang, Peijun Qing, Ming Cheng, Soroush Vosoughi, and Jiang Gui. 2025.
\newblock \href {https://arxiv.org/abs/2502.06020} {Temporal working memory: Query-guided segment refinement for enhanced multimodal understanding}.

\bibitem[{Friedman et~al.(2018)Friedman, Johnson, and Williams}]{friedman2018long}
Gary~N Friedman, Luke Johnson, and Zachary~M Williams. 2018.
\newblock \href {https://doi.org/10.3389/fpsyg.2018.01896} {Long-term visual memory and its role in learning suppression}.
\newblock \emph{Frontiers in Psychology}, 9:1896.

\bibitem[{Gao et~al.(2024)Gao, Ding, Cui, Zhao, Wang, Liu, and Qin}]{gao2024selfevolvinggptlifelongautonomous}
Jinglong Gao, Xiao Ding, Yiming Cui, Jianbai Zhao, Hepeng Wang, Ting Liu, and Bing Qin. 2024.
\newblock \href {https://arxiv.org/abs/2407.08937} {Self-evolving gpt: A lifelong autonomous experiential learner}.

\bibitem[{Gu et~al.(2024)Gu, Kuwajerwala, Morin, Jatavallabhula, Sen, Agarwal, Rivera, Paul, Ellis, Chellappa et~al.}]{gu2024conceptgraphs}
Qiao Gu, Ali Kuwajerwala, Sacha Morin, Krishna~Murthy Jatavallabhula, Bipasha Sen, Aditya Agarwal, Corban Rivera, William Paul, Kirsty Ellis, Rama Chellappa, et~al. 2024.
\newblock Conceptgraphs: Open-vocabulary 3d scene graphs for perception and planning.
\newblock In \emph{2024 IEEE International Conference on Robotics and Automation (ICRA)}, pages 5021--5028. IEEE.

\bibitem[{Guo et~al.(2023)Guo, Zhang, Zhu, Tang, Ma, Han, Chen, Gao, Li, Li, and Heng}]{guo2023pointbindpointllmaligning}
Ziyu Guo, Renrui Zhang, Xiangyang Zhu, Yiwen Tang, Xianzheng Ma, Jiaming Han, Kexin Chen, Peng Gao, Xianzhi Li, Hongsheng Li, and Pheng-Ann Heng. 2023.
\newblock \href {https://arxiv.org/abs/2309.00615} {Point-bind \& point-llm: Aligning point cloud with multi-modality for 3d understanding, generation, and instruction following}.

\bibitem[{Hong et~al.(2023{\natexlab{a}})Hong, Lin, Du, Chen, Tenenbaum, and Gan}]{Hong_2023_CVPR}
Yining Hong, Chunru Lin, Yilun Du, Zhenfang Chen, Joshua~B. Tenenbaum, and Chuang Gan. 2023{\natexlab{a}}.
\newblock \href {https://doi.org/10.1109/CVPR52729.2023.00888} {3d concept learning and reasoning from multi-view images}.
\newblock In \emph{{IEEE/CVF} Conference on Computer Vision and Pattern Recognition, {CVPR} 2023, Vancouver, BC, Canada, June 17-24, 2023}, pages 9202--9212. {IEEE}.

\bibitem[{Hong et~al.(2023{\natexlab{b}})Hong, Zhen, Chen, Zheng, Du, Chen, and Gan}]{3dllm}
Yining Hong, Haoyu Zhen, Peihao Chen, Shuhong Zheng, Yilun Du, Zhenfang Chen, and Chuang Gan. 2023{\natexlab{b}}.
\newblock \href {http://papers.nips.cc/paper\_files/paper/2023/hash/413885e70482b95dcbeeddc1daf39177-Abstract-Conference.html} {3d-llm: Injecting the 3d world into large language models}.
\newblock In \emph{Advances in Neural Information Processing Systems 36: Annual Conference on Neural Information Processing Systems 2023, NeurIPS 2023, New Orleans, LA, USA, December 10 - 16, 2023}.

\bibitem[{Hong et~al.(2024)Hong, Zheng, Chen, Wang, Li, and Gan}]{Hong_2024_CVPR}
Yining Hong, Zishuo Zheng, Peihao Chen, Yian Wang, Junyan Li, and Chuang Gan. 2024.
\newblock \href {https://doi.org/10.1109/CVPR52733.2024.02494} {Multiply: {A} multisensory object-centric embodied large language model in 3d world}.
\newblock In \emph{{IEEE/CVF} Conference on Computer Vision and Pattern Recognition, {CVPR} 2024, Seattle, WA, USA, June 16-22, 2024}, pages 26396--26406. {IEEE}.

\bibitem[{Huang et~al.(2024{\natexlab{a}})Huang, Chen, Wang, Huang, Xu, Wang, Liu, Cheng, Zhao, Pang, and Zhao}]{huang2024chat}
Haifeng Huang, Yilun Chen, Zehan Wang, Rongjie Huang, Runsen Xu, Tai Wang, Luping Liu, Xize Cheng, Yang Zhao, Jiangmiao Pang, and Zhou Zhao. 2024{\natexlab{a}}.
\newblock \href {http://papers.nips.cc/paper\_files/paper/2024/hash/cebbd24f1e50bcb63d015611fe0fe767-Abstract-Conference.html} {Chat-scene: Bridging 3d scene and large language models with object identifiers}.
\newblock In \emph{Advances in Neural Information Processing Systems 38: Annual Conference on Neural Information Processing Systems 2024, NeurIPS 2024, Vancouver, BC, Canada, December 10 - 15, 2024}.

\bibitem[{Huang et~al.(2024{\natexlab{b}})Huang, Yong, Ma, Linghu, Li, Wang, Li, Zhu, Jia, and Huang}]{huang2024embodied}
Jiangyong Huang, Silong Yong, Xiaojian Ma, Xiongkun Linghu, Puhao Li, Yan Wang, Qing Li, Song{-}Chun Zhu, Baoxiong Jia, and Siyuan Huang. 2024{\natexlab{b}}.
\newblock \href {https://openreview.net/forum?id=V4qV08Vk6S} {An embodied generalist agent in 3d world}.
\newblock In \emph{Forty-first International Conference on Machine Learning, {ICML} 2024, Vienna, Austria, July 21-27, 2024}. OpenReview.net.

\bibitem[{Intelligence et~al.(2025)Intelligence, Black, Brown, Darpinian, Dhabalia, Driess, Esmail, Equi, Finn, Fusai, Galliker, Ghosh, Groom, Hausman, Ichter, Jakubczak, Jones, Ke, LeBlanc, Levine, Li-Bell, Mothukuri, Nair, Pertsch, Ren, Shi, Smith, Springenberg, Stachowicz, Tanner, Vuong, Walke, Walling, Wang, Yu, and Zhilinsky}]{intelligence2025pi05visionlanguageactionmodelopenworld}
Physical Intelligence, Kevin Black, Noah Brown, James Darpinian, Karan Dhabalia, Danny Driess, Adnan Esmail, Michael Equi, Chelsea Finn, Niccolo Fusai, Manuel~Y. Galliker, Dibya Ghosh, Lachy Groom, Karol Hausman, Brian Ichter, Szymon Jakubczak, Tim Jones, Liyiming Ke, Devin LeBlanc, Sergey Levine, Adrian Li-Bell, Mohith Mothukuri, Suraj Nair, Karl Pertsch, Allen~Z. Ren, Lucy~Xiaoyang Shi, Laura Smith, Jost~Tobias Springenberg, Kyle Stachowicz, James Tanner, Quan Vuong, Homer Walke, Anna Walling, Haohuan Wang, Lili Yu, and Ury Zhilinsky. 2025.
\newblock \href {https://arxiv.org/abs/2504.16054} {$\pi_{0.5}$: a vision-language-action model with open-world generalization}.

\bibitem[{Khanna et~al.(2024)Khanna, Ramrakhya, Chhablani, Yenamandra, Gervet, Chang, Kira, Chaplot, Batra, and Mottaghi}]{khanna2024goatbench}
Mukul Khanna, Ram Ramrakhya, Gunjan Chhablani, Sriram Yenamandra, Th{\'{e}}ophile Gervet, Matthew Chang, Zsolt Kira, Devendra~Singh Chaplot, Dhruv Batra, and Roozbeh Mottaghi. 2024.
\newblock \href {https://doi.org/10.1109/CVPR52733.2024.01549} {Goat-bench: {A} benchmark for multi-modal lifelong navigation}.
\newblock In \emph{{IEEE/CVF} Conference on Computer Vision and Pattern Recognition, {CVPR} 2024, Seattle, WA, USA, June 16-22, 2024}, pages 16373--16383. {IEEE}.

\bibitem[{Kolve et~al.(2017)Kolve, Mottaghi, Han, VanderBilt, Weihs, Herrasti, Gordon, Zhu, Gupta, and Farhadi}]{ai2thor}
Eric Kolve, Roozbeh Mottaghi, Winson Han, Eli VanderBilt, Luca Weihs, Alvaro Herrasti, Daniel Gordon, Yuke Zhu, Abhinav Gupta, and Ali Farhadi. 2017.
\newblock {AI2-THOR: An Interactive 3D Environment for Visual AI}.
\newblock \emph{arXiv}.

\bibitem[{Krantz et~al.(2022)Krantz, Lee, Malik, Batra, and Chaplot}]{krantz2022instancespecificimagegoalnavigation}
Jacob Krantz, Stefan Lee, Jitendra Malik, Dhruv Batra, and Devendra~Singh Chaplot. 2022.
\newblock \href {https://arxiv.org/abs/2211.15876} {Instance-specific image goal navigation: Training embodied agents to find object instances}.

\bibitem[{Li et~al.(2024{\natexlab{a}})Li, Zhang, Wong, Gokmen, Srivastava, Martín-Martín, Wang, Levine, Ai, Martinez, Yin, Lingelbach, Hwang, Hiranaka, Garlanka, Aydin, Lee, Sun, Anvari, Sharma, Bansal, Hunter, Kim, Lou, Matthews, Villa-Renteria, Tang, Tang, Xia, Li, Savarese, Gweon, Liu, Wu, and Fei-Fei}]{li2024behavior1khumancenteredembodiedai}
Chengshu Li, Ruohan Zhang, Josiah Wong, Cem Gokmen, Sanjana Srivastava, Roberto Martín-Martín, Chen Wang, Gabrael Levine, Wensi Ai, Benjamin Martinez, Hang Yin, Michael Lingelbach, Minjune Hwang, Ayano Hiranaka, Sujay Garlanka, Arman Aydin, Sharon Lee, Jiankai Sun, Mona Anvari, Manasi Sharma, Dhruva Bansal, Samuel Hunter, Kyu-Young Kim, Alan Lou, Caleb~R Matthews, Ivan Villa-Renteria, Jerry~Huayang Tang, Claire Tang, Fei Xia, Yunzhu Li, Silvio Savarese, Hyowon Gweon, C.~Karen Liu, Jiajun Wu, and Li~Fei-Fei. 2024{\natexlab{a}}.
\newblock \href {https://arxiv.org/abs/2403.09227} {Behavior-1k: A human-centered, embodied ai benchmark with 1,000 everyday activities and realistic simulation}.

\bibitem[{Li et~al.(2024{\natexlab{b}})Li, Zhao, Wang, Wang, Zhou, Srivastava, Gokmen, Lee, Li, Zhang, Liu, Liang, Fei{-}Fei, Mao, and Wu}]{li2024embodied}
Manling Li, Shiyu Zhao, Qineng Wang, Kangrui Wang, Yu~Zhou, Sanjana Srivastava, Cem Gokmen, Tony Lee, Li~Erran Li, Ruohan Zhang, Weiyu Liu, Percy Liang, Li~Fei{-}Fei, Jiayuan Mao, and Jiajun Wu. 2024{\natexlab{b}}.
\newblock \href {http://papers.nips.cc/paper\_files/paper/2024/hash/b631da756d1573c24c9ba9c702fde5a9-Abstract-Datasets\_and\_Benchmarks\_Track.html} {Embodied agent interface: Benchmarking llms for embodied decision making}.
\newblock In \emph{Advances in Neural Information Processing Systems 38: Annual Conference on Neural Information Processing Systems 2024, NeurIPS 2024, Vancouver, BC, Canada, December 10 - 15, 2024}.

\bibitem[{Liu et~al.(2024)Liu, Zhang, Gu, Iong, Xu, Song, Zhang, Lai, Liu, Zhao et~al.}]{liu2024visualagentbench}
Xiao Liu, Tianjie Zhang, Yu~Gu, Iat~Long Iong, Yifan Xu, Xixuan Song, Shudan Zhang, Hanyu Lai, Xinyi Liu, Hanlin Zhao, et~al. 2024.
\newblock \href {https://arxiv.org/abs/2408.06327} {Visualagentbench: Towards large multimodal models as visual foundation agents}.
\newblock \emph{ArXiv preprint}, abs/2408.06327.

\bibitem[{Majumdar et~al.(2024)Majumdar, Ajay, Zhang, Putta, Yenamandra, Henaff, Silwal, McVay, Maksymets, Arnaud, Yadav, Li, Newman, Sharma, Berges, Zhang, Agrawal, Bisk, Batra, Kalakrishnan, Meier, Paxton, Sax, and Rajeswaran}]{OpenEQA2023}
Arjun Majumdar, Anurag Ajay, Xiaohan Zhang, Pranav Putta, Sriram Yenamandra, Mikael Henaff, Sneha Silwal, Paul McVay, Oleksandr Maksymets, Sergio Arnaud, Karmesh Yadav, Qiyang Li, Ben Newman, Mohit Sharma, Vincent{-}Pierre Berges, Shiqi Zhang, Pulkit Agrawal, Yonatan Bisk, Dhruv Batra, Mrinal Kalakrishnan, Franziska Meier, Chris Paxton, Alexander Sax, and Aravind Rajeswaran. 2024.
\newblock \href {https://doi.org/10.1109/CVPR52733.2024.01560} {Openeqa: Embodied question answering in the era of foundation models}.
\newblock In \emph{{IEEE/CVF} Conference on Computer Vision and Pattern Recognition, {CVPR} 2024, Seattle, WA, USA, June 16-22, 2024}, pages 16488--16498. {IEEE}.

\bibitem[{Packer et~al.(2023)Packer, Wooders, Lin, Fang, Patil, Stoica, and Gonzalez}]{packer2024memgptllmsoperatingsystems}
Charles Packer, Sarah Wooders, Kevin Lin, Vivian Fang, Shishir~G. Patil, Ion Stoica, and Joseph~E. Gonzalez. 2023.
\newblock \href {https://arxiv.org/abs/2310.08560} {Memgpt: Towards llms as operating systems}.

\bibitem[{Paszke et~al.(2019)Paszke, Gross, Massa, Lerer, Bradbury, Chanan, Killeen, Lin, Gribonval, Jozefowicz et~al.}]{pytorch}
Adam Paszke, Sam Gross, Francisco Massa, Adam Lerer, James Bradbury, Gregory Chanan, Trevor Killeen, Zeming Lin, Florian Gribonval, Rafal Jozefowicz, et~al. 2019.
\newblock Pytorch.
\newblock \url{https://pytorch.org/}.

\bibitem[{Ramakrishnan et~al.(2021)Ramakrishnan, Gokaslan, Wijmans, Maksymets, Clegg, Turner, Undersander, Galuba, Westbury, Chang, Savva, Zhao, and Batra}]{ramakrishnan2021hm3d}
Santhosh~Kumar Ramakrishnan, Aaron Gokaslan, Erik Wijmans, Oleksandr Maksymets, Alexander Clegg, John~M Turner, Eric Undersander, Wojciech Galuba, Andrew Westbury, Angel~X Chang, Manolis Savva, Yili Zhao, and Dhruv Batra. 2021.
\newblock \href {https://arxiv.org/abs/2109.08238} {Habitat-matterport 3d dataset ({HM}3d): 1000 large-scale 3d environments for embodied {AI}}.
\newblock volume abs/2109.08238.

\bibitem[{Shen et~al.(2021)Shen, Xia, Li, Mart\'in-Mart\'in, Fan, Wang, P\'erez-D'Arpino, Buch, Srivastava, Tchapmi, Tchapmi, Vainio, Wong, Fei-Fei, and Savarese}]{shen2021igibson}
Bokui Shen, Fei Xia, Chengshu Li, Roberto Mart\'in-Mart\'in, Linxi Fan, Guanzhi Wang, Claudia P\'erez-D'Arpino, Shyamal Buch, Sanjana Srivastava, Lyne~P. Tchapmi, Micael~E. Tchapmi, Kent Vainio, Josiah Wong, Li~Fei-Fei, and Silvio Savarese. 2021.
\newblock igibson 1.0: a simulation environment for interactive tasks in large realistic scenes.
\newblock In \emph{2021 IEEE/RSJ International Conference on Intelligent Robots and Systems (IROS)}, page accepted. IEEE.

\bibitem[{Shinn et~al.(2023)Shinn, Cassano, Gopinath, Narasimhan, and Yao}]{shinn2023reflexion}
Noah Shinn, Federico Cassano, Ashwin Gopinath, Karthik Narasimhan, and Shunyu Yao. 2023.
\newblock \href {http://papers.nips.cc/paper\_files/paper/2023/hash/1b44b878bb782e6954cd888628510e90-Abstract-Conference.html} {Reflexion: language agents with verbal reinforcement learning}.
\newblock In \emph{Advances in Neural Information Processing Systems 36: Annual Conference on Neural Information Processing Systems 2023, NeurIPS 2023, New Orleans, LA, USA, December 10 - 16, 2023}.

\bibitem[{Shridhar et~al.(2020)Shridhar, Thomason, Gordon, Bisk, Han, Mottaghi, Zettlemoyer, and Fox}]{Shridhar_2020_CVPR_alfred}
Mohit Shridhar, Jesse Thomason, Daniel Gordon, Yonatan Bisk, Winson Han, Roozbeh Mottaghi, Luke Zettlemoyer, and Dieter Fox. 2020.
\newblock \href {https://doi.org/10.1109/CVPR42600.2020.01075} {{ALFRED:} {A} benchmark for interpreting grounded instructions for everyday tasks}.
\newblock In \emph{2020 {IEEE/CVF} Conference on Computer Vision and Pattern Recognition, {CVPR} 2020, Seattle, WA, USA, June 13-19, 2020}, pages 10737--10746. {IEEE}.

\bibitem[{Shridhar et~al.(2021)Shridhar, Yuan, C{\^{o}}t{\'{e}}, Bisk, Trischler, and Hausknecht}]{ALFWorld20}
Mohit Shridhar, Xingdi Yuan, Marc{-}Alexandre C{\^{o}}t{\'{e}}, Yonatan Bisk, Adam Trischler, and Matthew~J. Hausknecht. 2021.
\newblock \href {https://openreview.net/forum?id=0IOX0YcCdTn} {Alfworld: Aligning text and embodied environments for interactive learning}.
\newblock In \emph{9th International Conference on Learning Representations, {ICLR} 2021, Virtual Event, Austria, May 3-7, 2021}. OpenReview.net.

\bibitem[{Szot et~al.(2021)Szot, Clegg, Undersander, Wijmans, Zhao, Turner, Maestre, Mukadam, Chaplot, Maksymets, Gokaslan, Vondrus, Dharur, Meier, Galuba, Chang, Kira, Koltun, Malik, Savva, and Batra}]{szot2021habitat}
Andrew Szot, Alexander Clegg, Eric Undersander, Erik Wijmans, Yili Zhao, John Turner, Noah Maestre, Mustafa Mukadam, Devendra~Singh Chaplot, Oleksandr Maksymets, Aaron Gokaslan, Vladimir Vondrus, Sameer Dharur, Franziska Meier, Wojciech Galuba, Angel~X. Chang, Zsolt Kira, Vladlen Koltun, Jitendra Malik, Manolis Savva, and Dhruv Batra. 2021.
\newblock \href {https://proceedings.neurips.cc/paper/2021/hash/021bbc7ee20b71134d53e20206bd6feb-Abstract.html} {Habitat 2.0: Training home assistants to rearrange their habitat}.
\newblock In \emph{Advances in Neural Information Processing Systems 34: Annual Conference on Neural Information Processing Systems 2021, NeurIPS 2021, December 6-14, 2021, virtual}, pages 251--266.

\bibitem[{Szot et~al.(2024)Szot, Schwarzer, Agrawal, Mazoure, Metcalf, Talbott, Mackraz, Hjelm, and Toshev}]{szot2023large}
Andrew Szot, Max Schwarzer, Harsh Agrawal, Bogdan Mazoure, Rin Metcalf, Walter Talbott, Natalie Mackraz, R.~Devon Hjelm, and Alexander~T. Toshev. 2024.
\newblock \href {https://openreview.net/forum?id=u6imHU4Ebu} {Large language models as generalizable policies for embodied tasks}.
\newblock In \emph{The Twelfth International Conference on Learning Representations, {ICLR} 2024, Vienna, Austria, May 7-11, 2024}. OpenReview.net.

\bibitem[{Team et~al.(2023)Team, Anil, Borgeaud, Wu, Alayrac, Yu, Soricut, Schalkwyk, Dai, Hauth et~al.}]{team2023gemini}
Gemini Team, Rohan Anil, Sebastian Borgeaud, Yonghui Wu, Jean-Baptiste Alayrac, Jiahui Yu, Radu Soricut, Johan Schalkwyk, Andrew~M Dai, Anja Hauth, et~al. 2023.
\newblock \href {https://arxiv.org/abs/2312.11805} {Gemini: a family of highly capable multimodal models}.
\newblock \emph{ArXiv preprint}, abs/2312.11805.

\bibitem[{team(2017--2025)}]{xla}
XLA team. 2017--2025.
\newblock Xla: Optimizing compiler for machine learning.
\newblock \url{https://www.tensorflow.org/xla}.

\bibitem[{Wang et~al.(2021)Wang, Wang, and Liu}]{wang2021temporal}
Hao Wang, Weining Wang, and Jing Liu. 2021.
\newblock Temporal memory attention for video semantic segmentation.
\newblock In \emph{2021 IEEE International Conference on Image Processing (ICIP)}, pages 2254--2258. IEEE.

\bibitem[{Wang and Agapito(2024)}]{wang20243d}
Hengyi Wang and Lourdes Agapito. 2024.
\newblock \href {https://arxiv.org/abs/2408.16061} {3d reconstruction with spatial memory}.
\newblock \emph{ArXiv preprint}, abs/2408.16061.

\bibitem[{Wang et~al.(2024)Wang, Yu, Zhao, Sun, Hou, Liang, Hu, Han, and Gan}]{wang2025karmaaugmentingembodiedai}
Zixuan Wang, Bo~Yu, Junzhe Zhao, Wenhao Sun, Sai Hou, Shuai Liang, Xing Hu, Yinhe Han, and Yiming Gan. 2024.
\newblock \href {https://arxiv.org/abs/2409.14908} {Karma: Augmenting embodied ai agents with long-and-short term memory systems}.

\bibitem[{Wijmans et~al.(2019)Wijmans, Datta, Maksymets, Das, Gkioxari, Lee, Essa, Parikh, and Batra}]{wijmans2019embodiedquestionansweringphotorealistic}
Erik Wijmans, Samyak Datta, Oleksandr Maksymets, Abhishek Das, Georgia Gkioxari, Stefan Lee, Irfan Essa, Devi Parikh, and Dhruv Batra. 2019.
\newblock \href {https://doi.org/10.1109/CVPR.2019.00682} {Embodied question answering in photorealistic environments with point cloud perception}.
\newblock In \emph{{IEEE} Conference on Computer Vision and Pattern Recognition, {CVPR} 2019, Long Beach, CA, USA, June 16-20, 2019}, pages 6659--6668. Computer Vision Foundation / {IEEE}.

\bibitem[{Xie et~al.(2024)Xie, Min, Ji, Yang, Zhang, Bajaj, Salakhutdinov, Johnson-Roberson, and Bisk}]{xie2024embodiedraggeneralnonparametricembodied}
Quanting Xie, So~Yeon Min, Pengliang Ji, Yue Yang, Tianyi Zhang, Aarav Bajaj, Ruslan Salakhutdinov, Matthew Johnson-Roberson, and Yonatan Bisk. 2024.
\newblock \href {https://arxiv.org/abs/2409.18313} {Embodied-rag: General non-parametric embodied memory for retrieval and generation}.

\bibitem[{Xu et~al.(2025{\natexlab{a}})Xu, Wang, Wang, Chen, Pang, and Lin}]{pointllm}
Runsen Xu, Xiaolong Wang, Tai Wang, Yilun Chen, Jiangmiao Pang, and Dahua Lin. 2025{\natexlab{a}}.
\newblock Pointllm: Empowering large language models to understand point clouds.
\newblock In \emph{Computer Vision -- ECCV 2024}, pages 131--147. Springer Nature Switzerland.

\bibitem[{Xu et~al.(2025{\natexlab{b}})Xu, Liang, Mei, Gao, Tan, and Zhang}]{xu2025mem}
Wujiang Xu, Zujie Liang, Kai Mei, Hang Gao, Juntao Tan, and Yongfeng Zhang. 2025{\natexlab{b}}.
\newblock \href {https://arxiv.org/abs/2502.12110} {A-mem: Agentic memory for llm agents}.
\newblock \emph{ArXiv preprint}, abs/2502.12110.

\bibitem[{Yang et~al.(2024)Yang, Yang, Gupta, Han, Fei-Fei, and Xie}]{yang2024think}
Jihan Yang, Shusheng Yang, Anjali~W. Gupta, Rilyn Han, Li~Fei-Fei, and Saining Xie. 2024.
\newblock \href {https://arxiv.org/abs/2412.14171} {{Thinking in Space: How Multimodal Large Language Models See, Remember and Recall Spaces}}.
\newblock \emph{ArXiv preprint}, abs/2412.14171.

\bibitem[{Yang et~al.(2025{\natexlab{a}})Yang, Chen, Zhang, Zhao, Qian, Wang, Wang, Koripella, Movahedi, Li, Ji, Zhang, and Zhang}]{yang2025embodiedbenchcomprehensivebenchmarkingmultimodal}
Rui Yang, Hanyang Chen, Junyu Zhang, Mark Zhao, Cheng Qian, Kangrui Wang, Qineng Wang, Teja~Venkat Koripella, Marziyeh Movahedi, Manling Li, Heng Ji, Huan Zhang, and Tong Zhang. 2025{\natexlab{a}}.
\newblock \href {https://arxiv.org/abs/2502.09560} {Embodiedbench: Comprehensive benchmarking multi-modal large language models for vision-driven embodied agents}.

\bibitem[{Yang et~al.(2025{\natexlab{b}})Yang, Yang, Zhou, Chen, Zhang, Du, and Gan}]{yang20243dmem3dscenememory}
Yuncong Yang, Han Yang, Jiachen Zhou, Peihao Chen, Hongxin Zhang, Yilun Du, and Chuang Gan. 2025{\natexlab{b}}.
\newblock 3d-mem: 3d scene memory for embodied exploration and reasoning.
\newblock In \emph{Proceedings of the IEEE/CVF Conference on Computer Vision and Pattern Recognition (CVPR)}.

\bibitem[{Yu et~al.(2019)Yu, Chen, Gkioxari, Bansal, Berg, and Batra}]{yu2019multitargetembodiedquestionanswering}
Licheng Yu, Xinlei Chen, Georgia Gkioxari, Mohit Bansal, Tamara~L. Berg, and Dhruv Batra. 2019.
\newblock \href {https://doi.org/10.1109/CVPR.2019.00647} {Multi-target embodied question answering}.
\newblock In \emph{{IEEE} Conference on Computer Vision and Pattern Recognition, {CVPR} 2019, Long Beach, CA, USA, June 16-20, 2019}, pages 6309--6318. Computer Vision Foundation / {IEEE}.

\bibitem[{Zhang et~al.(2023)Zhang, Chen, Zhang, Xu, Zhao, and Yu}]{DanyangZhang2023_Rememberer}
Danyang Zhang, Lu~Chen, Situo Zhang, Hongshen Xu, Zihan Zhao, and Kai Yu. 2023.
\newblock \href {https://arxiv.org/abs/2306.07929} {Large language model is semi-parametric reinforcement learning agent}.
\newblock \emph{ArXiv preprint}, abs/2306.07929.

\bibitem[{Zhang et~al.(2024)Zhang, Bo, Ma, Li, Chen, Dai, Zhu, Dong, and Wen}]{zhang2024surveymemorymechanismlarge}
Zeyu Zhang, Xiaohe Bo, Chen Ma, Rui Li, Xu~Chen, Quanyu Dai, Jieming Zhu, Zhenhua Dong, and Ji-Rong Wen. 2024.
\newblock \href {https://arxiv.org/abs/2404.13501} {A survey on the memory mechanism of large language model based agents}.

\bibitem[{Zhao et~al.(2024)Zhao, Huang, Xu, Lin, Liu, and Huang}]{zhao2024expel}
Andrew Zhao, Daniel Huang, Quentin Xu, Matthieu Lin, Yong{-}Jin Liu, and Gao Huang. 2024.
\newblock \href {https://doi.org/10.1609/AAAI.V38I17.29936} {Expel: {LLM} agents are experiential learners}.
\newblock In \emph{Thirty-Eighth {AAAI} Conference on Artificial Intelligence, {AAAI} 2024, Thirty-Sixth Conference on Innovative Applications of Artificial Intelligence, {IAAI} 2024, Fourteenth Symposium on Educational Advances in Artificial Intelligence, {EAAI} 2014, February 20-27, 2024, Vancouver, Canada}, pages 19632--19642. {AAAI} Press.

\bibitem[{Zhao et~al.(2025)Zhao, Lu, Kim, Fu, Zhang, Wu, Li, Ma, Han, Finn, Handa, Liu, Xiang, Wetzstein, and Lin}]{zhao2025cotvlavisualchainofthoughtreasoning}
Qingqing Zhao, Yao Lu, Moo~Jin Kim, Zipeng Fu, Zhuoyang Zhang, Yecheng Wu, Zhaoshuo Li, Qianli Ma, Song Han, Chelsea Finn, Ankur Handa, Ming-Yu Liu, Donglai Xiang, Gordon Wetzstein, and Tsung-Yi Lin. 2025.
\newblock \href {https://arxiv.org/abs/2503.22020} {Cot-vla: Visual chain-of-thought reasoning for vision-language-action models}.

\bibitem[{Zhen et~al.(2024)Zhen, Qiu, Chen, Yang, Yan, Du, Hong, and Gan}]{zhen20243dvla}
Haoyu Zhen, Xiaowen Qiu, Peihao Chen, Jincheng Yang, Xin Yan, Yilun Du, Yining Hong, and Chuang Gan. 2024.
\newblock \href {https://arxiv.org/abs/2403.09631} {3d-vla: 3d vision-language-action generative world model}.
\newblock \emph{ArXiv preprint}, abs/2403.09631.

\bibitem[{Zhou et~al.(2024)Zhou, Wang, Ma, Liu, Huang, and Wang}]{zhou2023uni3d}
Junsheng Zhou, Jinsheng Wang, Baorui Ma, Yu{-}Shen Liu, Tiejun Huang, and Xinlong Wang. 2024.
\newblock \href {https://openreview.net/forum?id=wcaE4Dfgt8} {Uni3d: Exploring unified 3d representation at scale}.
\newblock In \emph{The Twelfth International Conference on Learning Representations, {ICLR} 2024, Vienna, Austria, May 7-11, 2024}. OpenReview.net.

\bibitem[{Zhu et~al.(2024)Zhu, Wang, Zhang, Pang, and Liu}]{zhu2024llava}
Chenming Zhu, Tai Wang, Wenwei Zhang, Jiangmiao Pang, and Xihui Liu. 2024.
\newblock \href {https://arxiv.org/abs/2409.18125} {Llava-3d: A simple yet effective pathway to empowering lmms with 3d-awareness}.
\newblock \emph{ArXiv preprint}, abs/2409.18125.

\bibitem[{Zlotnik and Vansintjan(2019)}]{zlotnik2019memory}
Guillermo Zlotnik and Aaron Vansintjan. 2019.
\newblock \href {https://doi.org/10.3389/fpsyg.2019.02523} {Memory: An extended definition}.
\newblock \emph{Frontiers in Psychology}, 10:2523.

\bibitem[{Zou et~al.(2025)Zou, Song, Qiu, Peng, Ye, Liu, and Wang}]{m3}
Xueyan Zou, Yuchen Song, Ri-Zhao Qiu, Xuanbin Peng, Jianglong Ye, Sifei Liu, and Xiaolong Wang. 2025.
\newblock M3: 3d-spatial multimodal memory.
\newblock In \emph{ICLR}.

\end{thebibliography}
\bibliographystyle{natbib}

%%%%%%%%%%%%%%%%%%%%%%%%%%%%%%%%%%%%%%%%%%%%%%%%%%%%%%%%%%%%

\clearpage
%%%%%%%%%%%%%%%%%%%%%%%%%%%%%%%%%%%%%%%%%%%%%%%%%%%%%%%%%%%%

\newpage
\section*{NeurIPS Paper Checklist}

\begin{enumerate}

\item {\bf Claims}
    \item[] Question: Do the main claims made in the abstract and introduction accurately reflect the paper's contributions and scope?
    \item[] Answer:\answerYes{}
    \item[] Justification: We clearly state our main claims in abstract and introduction.

\item {\bf Limitations}
    \item[] Question: Does the paper discuss the limitations of the work performed by the authors?
    \item[] Answer: \answerYes{}
    \item[] Justification: We discuss the limitations in Section~\ref{sec: conclusion}.

\item {\bf Theory assumptions and proofs}
    \item[] Question: For each theoretical result, does the paper provide the full set of assumptions and a complete (and correct) proof?
    \item[] Answer: \answerNA{}.
    \item[] Justification: This paper doesn’t introduce new theorems.

    \item {\bf Experimental result reproducibility}
    \item[] Question: Does the paper fully disclose all the information needed to reproduce the main experimental results of the paper to the extent that it affects the main claims and/or conclusions of the paper (regardless of whether the code and data are provided or not)?
    \item[] Answer: \answerYes{}
    \item[] Justification:  Yes, we fully disclose all the information, please refer to Section~\ref{sec: method} and our experimental setup in Section~\ref{sec: experimental setup}
    \item[] Guidelines:
    \begin{itemize}
        \item The answer NA means that the paper does not include experiments.
        \item If the paper includes experiments, a No answer to this question will not be perceived well by the reviewers: Making the paper reproducible is important, regardless of whether the code and data are provided or not.
        \item If the contribution is a dataset and/or model, the authors should describe the steps taken to make their results reproducible or verifiable. 
        \item Depending on the contribution, reproducibility can be accomplished in various ways. For example, if the contribution is a novel architecture, describing the architecture fully might suffice, or if the contribution is a specific model and empirical evaluation, it may be necessary to either make it possible for others to replicate the model with the ssame dataset, or provide access to the model. In general. releasing code and data is often one good way to accomplish this, but reproducibility can also be provided via detailed instructions for how to replicate the results, access to a hosted model (e.g., in the case of a large language model), releasing of a model checkpoint, or other means that are appropriate to the research performed.
        \item While NeurIPS does not require releasing code, the conference does require all submissions to provide some reasonable avenue for reproducibility, which may depend on the nature of the contribution. For example
        \begin{enumerate}
            \item If the contribution is primarily a new algorithm, the paper should make it clear how to reproduce that algorithm.
            \item If the contribution is primarily a new model architecture, the paper should describe the architecture clearly and fully.
            \item If the contribution is a new model (e.g., a large language model), then there should either be a way to access this model for reproducing the results or a way to reproduce the model (e.g., with an open-source dataset or instructions for how to construct the dataset).
            \item We recognize that reproducibility may be tricky in some cases, in which case authors are welcome to describe the particular way they provide for reproducibility. In the case of closed-source models, it may be that access to the model is limited in some way (e.g., to registered users), but it should be possible for other researchers to have some path to reproducing or verifying the results.
        \end{enumerate}
    \end{itemize}

\item {\bf Open access to data and code}
    \item[] Question: Does the paper provide open access to the data and code, with sufficient instructions to faithfully reproduce the main experimental results, as described in supplemental material?
    \item[] Answer:  \answerYes{}
    \item[] Justification: We will release our code and data publicly after the review process. We also provide data sample and example code in the submitted supplemental material. 

\item {\bf Experimental setting/details}
    \item[] Question: Does the paper specify all the training and test details (e.g., data splits, hyperparameters, how they were chosen, type of optimizer, etc.) necessary to understand the results?
    \item[] Answer: \answerYes{}
    \item[] Justification: Please refer to our implementation details in Section~\ref{sec: experimental setup} and Appendix~\ref{appendix: implement details}.

\item {\bf Experiment statistical significance}
    \item[] Question: Does the paper report error bars suitably and correctly defined or other appropriate information about the statistical significance of the experiments?
    \item[] Answer: \answerNo{}
    \item[] Justification: We conduct experiments on Google TPU and the training of long-term memory 3D-LLM is expensive, we don't have the resources to run the experiments multiple times and calculate the error bar. 

\item {\bf Experiments compute resources}
    \item[] Question: For each experiment, does the paper provide sufficient information on the computer resources (type of compute workers, memory, time of execution) needed to reproduce the experiments?
    \item[] Answer: \answerYes{}
    \item[] Justification: Please refer to our implementation details in Section~\ref{sec: experimental setup} and Appendix~\ref{appendix: implement details}.
    
\item {\bf Code of ethics}
    \item[] Question: Does the research conducted in the paper conform, in every respect, with the NeurIPS Code of Ethics \url{https://neurips.cc/public/EthicsGuidelines}?
    \item[] Answer: \answerYes{}
    \item[] Justification: We discuss ethics concern and broader impact in Appendix~\ref{appendix: broader impact}.

\item {\bf Broader impacts}
    \item[] Question: Does the paper discuss both potential positive societal impacts and negative societal impacts of the work performed?
    \item[] Answer: \answerYes{}
    \item[] Justification: We discuss broader impact in Appendix~\ref{appendix: broader impact}.
    
\item {\bf Safeguards}
    \item[] Question: Does the paper describe safeguards that have been put in place for responsible release of data or models that have a high risk for misuse (e.g., pretrained language models, image generators, or scraped datasets)?
    \item[] Answer: \answerYes{}
    \item[] Justification: We require users to follow the guidelines such as Llama’s guidelines when release the model.

\item {\bf Licenses for existing assets}
    \item[] Question: Are the creators or original owners of assets (e.g., code, data, models), used in the paper, properly credited and are the license and terms of use explicitly mentioned and properly respected?
    \item[] Answer: \answerYes{}
    \item[] Justification: We properly credited the original owners and followed their license.

\item {\bf New assets}
    \item[] Question: Are new assets introduced in the paper well documented and is the documentation provided alongside the assets?
    \item[] Answer: \answerYes{}. 
    \item[] Justification: We document our new introduced benchmark in Section~\ref{sec:benchmark intro} and Section~\ref{sec:data collection}. We also provide the documentation in our dataset along with the data sample in the supplementary material. 

\item {\bf Crowdsourcing and research with human subjects}
    \item[] Question: For crowdsourcing experiments and research with human subjects, does the paper include the full text of instructions given to participants and screenshots, if applicable, as well as details about compensation (if any)? 
    \item[] Answer: \answerNA{}
    \item[] Justification: We authors conducted the human validation with clear instructions, no other research with external human subjects is performed. 

\item {\bf Institutional review board (IRB) approvals or equivalent for research with human subjects}
    \item[] Question: Does the paper describe potential risks incurred by study participants, whether such risks were disclosed to the subjects, and whether Institutional Review Board (IRB) approvals (or an equivalent approval/review based on the requirements of your country or institution) were obtained?
    \item[] Answer: \answerNA{}
    \item[] Justification: This paper does not involve crowdsourcing nor research with human subjects.

\item {\bf Declaration of LLM usage}
    \item[] Question: Does the paper describe the usage of LLMs if it is an important, original, or non-standard component of the core methods in this research? Note that if the LLM is used only for writing, editing, or formatting purposes and does not impact the core methodology, scientific rigorousness, or originality of the research, declaration is not required.
    %this research? 
    \item[] Answer: \answerNA{}.
    \item[] Justification: We only used LLMs for improving grammar in paper writing. 

\end{enumerate}

\appendix

\section{Broader Impact}
\label{appendix: broader impact}

The deployment and release of \model carry both potential benefits and risks. These considerations include visual aspects as well as common issues found in existing LLMs like Alpaca and Vicuna. Since \model is built on LLaMA, Vicuna, and CLIP, it inherits certain challenges associated with LLMs and vision encoders. 
Below, we outline the risks and the mitigation strategies implemented for the release of this model.

\paragraph{Hallucination}  Similar to other LLMs, \model might produce outputs that are not based on factual information or input data. This raises concerns about the accuracy of inferences, particularly in critical applications such as medical fields.

\paragraph{Biases} Biases present in the base models can be brought to \model, stemming from both the vision encoder (CLIP) and the language decoder (LLaMA / Vicuna). This may result in biased outcomes or unfair representations of diverse content.

\paragraph{Energy Consumption} We train our model on our training data split which contains about 26K trajectories. The training time only takes less than one day, which makes energy consumption not a primary concern.

\section{Environment Construction}
\label{appendix: env construction}

To support navigation‑centric interaction, the agent requires precise knowledge of two things: the traversable layout of each scene and the exact locations of all movable objects. Following 3D‑CLR~\citep{Hong_2023_CVPR}, we build this spatial substrate from HM3D’s richly annotated indoor scans. We rely on the semantic surface mesh that accompanies each scene to calculate the room and objects' locations. The mesh scan has a unique (24‑bit) hexadecimal color for every surface triangle that provides a semantic table that links each color to a surface label (e.g., floor or ceiling) and a room identifier.

We first derive an axis‑aligned bounding box for every room in each HM3D scene. We query the semantic table to retrieve, for every room identifier, the hex colors assigned to its floor and ceiling surfaces. Before processing individual rooms we aggregate the minimum heights of all floor surfaces in the entire scan, producing a global set of candidate floor elevations. For each room we then load the point clouds of its floor and ceiling. When both are available, the vertical bounds are given by the floor’s lowest point and the ceiling’s highest point. If the floor is missing, the lower bound is set to the highest global floor elevation that still lies below the room’s ceiling; if the ceiling is missing, the upper bound defaults to the highest point in the available cloud (i.e., the surface immediately above the floor). Rooms for which both surfaces are absent are discarded because no reliable vertical evidence is present. Horizontal limits are simply the minimum and maximum coordinates of the room’s floor and ceiling points, and together these six coordinates constitute the room’s axis‑aligned bounding box.

Based on the bounding boxes for each room, we calculate the bounding boxes of objects in that room by gathering the corresponding vertices and computing the bounding boxes of the object by taking the minimum and maximum coordinates along the three axes. We merge room‑level and object‑level bounding boxes by their shared room index. After the aforementioned processing, the dataset comprises 182 distinct 3‑D scenes covering a total of 2,602 rooms. 

\section{Embodied Task Data Examples}
\label{appendix: data examples}

In this section, we provide complete embodied task trajectories.

Here is an example of in-the-wild task: \\
\textbf{Task}: 
    Rearrange desk items by size. The items to be rearranged are: vase, box, and printer. \\
\textbf{Trajectory}:  \\
    "I'm in room 10. I will start exploring the room for the items I need.",
            "I see a printer(0) on the desk(0). I also see a box(0) on the cabinet(0).",
            "I need to remember their locations and sizes for comparison later.", \\
            "<GO TO NEW ROOM>",\\
            "Now I'm in room 11. I'm exploring this room for the items.",
            "I don't see a vase in the room, but I see a printer(0) on a rack(1).",
            "I will remember these objects and their locations.",\\
            "<GO TO NEW ROOM>",\\
            "Now I'm in room 6. I'm exploring this room.",
            "I do not see any of the items I am looking for.",\\
            "<GO TO NEW ROOM>",\\
            "I'm in room 8. I am exploring this room.",
            "I see flower vase(0) on the floor. I will remember the object and its location.",\\
            "<GO TO NEW ROOM>",\\
            "I'm in room 7. I am exploring this room.",
            "I do not see any of the items I am looking for.",\\
            "<GO TO NEW ROOM>",\\
            "I'm in room 9. I am exploring this room.",
            "I do not see any of the items I am looking for.",\\
            "<GO TO NEW ROOM>",\\
            "Now I am in room 5. I am exploring this room.",
            "I do not see any of the items I am looking for.",\\
            "<GO TO NEW ROOM>",\\
            "Now I am in room 2. I am exploring this room.",
            "I see a box(0) on the armchair(1). I will remember these objects and their locations.",\\
            "<GO TO NEW ROOM>",\\
            "I'm in room 1. I do not see any items that are part of the rearrangement.",\\
            "<GO TO NEW ROOM>",\\
            "Now I'm in room 12. I will start exploring this room.",
            "I think I've seen the vase, printer, and box in other rooms, so I'll head back to room 8 to pick up the vase first.",\\
            "<GO TO ROOM(8)>",\\
            "I need to pick up flower vase(0) in this room first. I pick up flower vase(0).",\\
            "<PICK UP flower vase(0) from room(8) in room(8)>",\\
            "I'm carrying flower vase(0) to room 10 where I saw printer and box.",\\
            "<GO TO ROOM(10)>",\\
            "Now I am in room 10. I will put flower vase(0) on the desk(0) for now so I can pick up box(0).",\\
            "<PUT DOWN flower vase(0) from room(8) on desk(0) in room(10)>",\\
            "Now I will pick up box(0) from cabinet(0).",\\
            "<PICK UP box(0) from room(10) in room(10)>",\\
            "Now I will put box(0) on the desk(0) so I can pick up printer(0).",\\
            "<PUT DOWN box(0) from room(10) on desk(0) in room(10)>",\\
            "Now I will pick up printer(0) from beside the desk(0).",\\
            "<PICK UP printer(0) from room(10) in room(10)>",\\
            "I need to compare the sizes of the vase, box and printer now to determine the rearranging order.",
            "I recall that printer(0) in this room had the largest bounding box dimensions based on my earlier observation. I need to put down printer(0).",\\
            "<PUT DOWN printer(0) from room(10) on desk(0) in room(10)>",\\
            "Now I will pick up box(0).",\\
            "<PICK UP box(0) from room(10) in room(10)>",\\
            "I remember that box(0) from cabinet(0) had medium bounding box dimensions, so I will put it on the desk.",\\
            "<PUT DOWN box(0) from room(10) on desk(0) in room(10)>",\\
            "Now I will pick up flower vase(0).",\\
            "<PICK UP flower vase(0) from room(8) in room(10)>",\\
            "I remember that flower vase(0) had the smallest bounding box dimensions, so I will put it on the desk.",\\
            "<PUT DOWN flower vase(0) from room(8) on desk(0) in room(10)>",\\
            "I will now arrange the objects from small to large: flower vase(0), box(0) and printer(0) on the desk(0).",
            "Now I pick up the medium object, box(0).",\\
            "<PICK UP box(0) from room(10) in room(10)>",\\
            "I will put the medium object on the left of the largest object, printer(0).",\\
            "<PUT DOWN box(0) from room(10) on desk(0) in room(10)>",\\
            "Now I pick up the smallest object, flower vase(0).",\\
            "<PICK UP flower vase(0) from room(8) in room(10)>",\\
            "I will put the smallest object on the left of the medium object, box(0).",\\
            "<PUT DOWN flower vase(0) from room(8) on desk(0) in room(10)>",\\
            "The objects are now arranged from small to large: flower vase(0), box(0) and printer(0) on the desk(0). Task Complete."

Here is an example of in-domain task:

\textbf{Task}:
Preparing for Cooking Stir-fried Tomato and Scrambled Eggs. You need to prepare the ingredients, seasonings and cooking tools. 

\textbf{Trajectory}:
  "I'm at room 4. It seems to be living room, I need to prepare ingredients and seasonings and cooking tools and setup them in the kitchen",
            "There is seasonings on the table.",\\
            "<PICK UP seasonings(0) from room(4) in room(4)>",\\
            "No other things found for my task",\\
            "<GO TO NEW ROOM>",\\
            "I'm at room 5. It seems to be dining room. I saw an apron on dining table.",
            "I need the apron but I have seasonings(0) on my hand. Lets find kitchen to put the seasonings first and go back for apron",\\
            "<GO TO NEW ROOM>",\\
            "I'm at room 8. It seems to be kitchen with two cabinets. I found tomatoes and eggs on the cabinet(1) that is far away from me and cooking pan in the sink.",
            "There are two countertops in this room. countertop(1) is closer to stove(0). I should setup on countertop(1).",\\
            "<PUT DOWN seasonings(0) from room(4) on countertop(1) in room(8)>",\\
            "<PICK UP tomatoes(0) from room(8) in room(8)>",\\
            "<PUT DOWN tomatoes(0) from room(8) on countertop(1) in room(8)>",\\
            "<PICK UP eggs(0) from room(8) in room(8)>",\\
            "<PUT DOWN eggs(0) from room(8) on countertop(1) in room(8)>",\\
            "<PICK UP cooking pan(0) from room(8) in room(8)>",\\
            "<PUT DOWN cooking pan(0) from room(8) on stove(0) in room(8)>",\\
            "I remember I saw apron in room 5. I need to find it back.",\\
            "<GO TO ROOM(5)>", \\
            "<PICK UP apron(0) from room(5) in room(5)>",\\
            "<GO TO ROOM(8)>",\\
            "<PUT DOWN apron(0) from room(5) on countertop(1) in room(8)>",\\
            "The setup in kitchen has been prepared. Task Complete."

\section{Implementation Details}
\label{appendix: implement details}
We implement our model based on LLaVA-3D~\citep{zhu2024llava}, modifying it to be compatible with Google TPUs with PyTorch/XLA frameworks~\citep{pytorch, xla} . We first expand the model’s context window to 8192 tokens to accommodate long-term memory inputs. We then fine-tune our proposed memory module along with the LLM decoder using our training split, initializing from LLaVA-3D’s pretrained weights. Training is conducted on 8 Google Cloud TPU v5p cores with a batch size of 256 for 1000 steps, which takes about 1 day to complete. We use Adam optimizer with learning rate of 2e-5 with no weight decay. Additionally, we apply a linear warmup of the learning rate during the initial 3\% steps, increasing from $10^{-8}$ to $10^{-5}$, followed by a cosine decay scheduler.

\section{Prompts for Gemini}
\label{appendix: generation prompts}

As mentioned in $\S$~\ref{sec:data collection}, we prompt Gemini to generate the long-term trajectories as illustrated in Table~\ref{tab:generate trajectory}, generate the question-answering tasks as shown in Table~\ref{tab:generate QA}, and generate caption tasks as shown in Table~\ref{tab:generate caption}. For open-ended QA evaluation, we followed standard LLM-as-judge protocol by prompting Gemini as illustrated in Table~\ref{tab: llm as judge}.

\begin{table*}[h!]\centering
\begin{minipage}{0.95\textwidth}   
\centering
\begin{tcolorbox} 
    \centering
    \small
    \begin{tabular}{p{0.95\textwidth}}
   { {\bf System message} } \\
    You are an AI assistant and task generator for a 3D embodied agent operating in a multi-room environment. The environment provides detailed object instance information, including bounding boxes and IDs. Your goal is to generate a complex task that requires the agent to explore multiple rooms, navigate, and crucially use long-term memory to recall details observed earlier. \\
    \midrule
    {\bf Prompt}  \\
    1. Environment and Object Information

    Object Representation:
    Each object is given with a bounding box in the format:
    ``<object\_name>(num)'': [x\_min, y\_min, z\_min], [x\_max, y\_max, z\_max]
    Here, (num) indicates the ID, with (0) being the closest to the origin [0,0,0]. IDs reset for each room (e.g., sofa(0) in room 2 and sofa(0) in room 4 if each room has one sofa).

    Actions Available:
        <GO TO ROOM(id)>: Navigate to a room that has already been visited.
        <GO TO NEW ROOM>: Navigate to a new, unexplored room (and unlock its objects). Do not use this for rooms that have been visited before.
        <PICK UP object\_name(id) from room(id) in room(id)>: Pick up an object that originally belongs to a specific room while in that same room.
        <PUT DOWN object\_name(id) from room(id) on object\_name(id) in room(id)>: Place an object (that originally belongs to a room) onto another object (such as a table or floor) in a room.

    New Objects:
    You can add extra objects to diversify the task. Important: Use only object names from the provided new\_objects\_name\_list. If a room already has an object with the same name, the new object should have a new ID (e.g., if lamp(0) exists, the added one should be lamp(1)). These extra objects are only for task design; the agent’s trajectory should not mention adding them.
\\\\
2. Task Design Requirements

    Multi-Room Exploration:
    Design a task that spans several rooms. The room order (given in a Room Order list) should be chosen so that necessary items are distributed across rooms. The agent should explore every room in the specified order.

    Long-Term Memory and Implicit Cues:
    Do not simply list all items as a checklist at the start. Instead:
        Provide a vague overall goal (e.g., “prepare a meal”).
        Later in the trajectory, have the agent recall these earlier observations when the need arises.
        Ensure the agent must remember something seen long ago rather than simply following an explicit list.

    Update Memory and make new decision based on your current observations:
        The agent originall planned to use one object for completing its task, but couldn't find it after exploration of rooms. It has to change to a another similar object to complete its task. 

    Inventory and Action Constraints:
        The agent can only hold one item at a time.
        Never perform consecutive PICK UP or PUT DOWN actions. If the agent holds an item, it must put it down before picking up another.
        When temporarily storing an object (e.g., on a table), include a “thought” explaining why the object is being set down and later recalled.
\\\\
    3.Reasoning and Object Comparisons:
    If your task requires choosing a specific object instance (e.g., selecting table(1) because it is bigger than table(0)), compare their bounding boxes and explain your choice in the trajectory.
    
\\
    For clarity, consider these examples: \textcolor[rgb]{0,0.7,0}{ \{In-context examples\} } \\
------------------- \\
    Here is the scene information: 
\textcolor[rgb]{0.8,0,0}{\{Input scene information\}} \\
\\

    \end{tabular}
\end{tcolorbox}
\caption{Prompt template for generating task trajectories. \textcolor[rgb]{0,0.7,0}{ \{In-context examples\} } are in-context examples. \textcolor[rgb]{0.8,0,0}{\{Input scene information\}} are scene, room and object semantics along with their bounding boxes.}
    \label{tab:generate trajectory}
\end{minipage}
\end{table*}

\begin{table*}[h!]\centering
\begin{minipage}{0.95\textwidth}   
\centering
\begin{tcolorbox} 
    \centering
    \small
    \begin{tabular}{p{0.95\textwidth}}
    {\bf Prompt}  \\
You are an AI assistant / task generator in the room. All object instances in 
this 3D scene are given, along with their bounding boxes and ids." 
Each object's bounding boxes are represented by a 3D coordinate '<obj\_name>(num)': [x min, y min, z min],[x max, y max, z max]' with units of meters, and each represents left-bottom corner and the right-top corner coordinate. \\\\
You will also receive a trajectory composed of the following tokens and reasoning chains.  \\

<GO TO ROOM(id)>: which navigates back to a specific room (id). This can only be done if the agent already go to this room. 
% <PICK UP object\_name(id) in room(id)>: pick up the object in a specific room (id).  
% <PUT DOWN object\_name(id) from room(id) on object\_name(id) in room(id)>: put down the object from a specific room (num) in a specific room (num) on a object\_name (id). 
 <PICK UP object\_name(id) from room(id) in room(id)>: Pick up an object that originally belongs to a specific room while in that same room.
<PUT DOWN object\_name(id) from room(id) on object\_name(id) in room(id)>: Place an object (that originally belongs to a room) onto another object (such as a table or floor) in a room.
<GO TO NEW ROOM>: which navigates to a new room you haven't explored and unlocks objects there. \\ \\

This trajectory is what the agent have executed over the past. You need to propose several questions and answers that focused on the reasoning abilities of the long-term memory of the agent. 
These reasoning questions should focus on what have changed temporally or spatially in this agent's memory. It's important that this change challenged the agent's memory.
For example the questions should contain object counting, spatial relation, comparison between objects across rooms, long-term multi-room room layout, long-term multi-room object navigation. 
Remember spatial memory is important, you should design questions that asked about the 3D object spatial relation and layout in the room that need the agent to perform a hard reasoning for the final answer. 
\\\\
    
    For clarity, consider these examples: \textcolor[rgb]{0,0.7,0}{ \{In-context examples\} } \\
------------------- \\
    Here is the scene information: 
\textcolor[rgb]{0.8,0,0}{\{Input scene information\}} \\
    Here is the agent's trajectory: 
\textcolor[rgb]{0,0,0.8}{\{Input agent's trajectory\}} \\
\\

    \end{tabular}
\end{tcolorbox}
\caption{Prompt template for generate QA data. \textcolor[rgb]{0,0.7,0}{ \{In-context examples\} } are in-context examples. \textcolor[rgb]{0.8,0,0}{\{Input scene information\}} are scene, room and object semantics along with their bounding boxes. \textcolor[rgb]{0,0,0.8}{\{Input agent's trajectory\}} is the 3D agent's explored trajectories and action chains.}
    \label{tab:generate QA}
\end{minipage}
\end{table*}

\begin{table*}[h!]\centering
\begin{minipage}{0.95\textwidth}   
\centering
\begin{tcolorbox} 
    \centering
    \small
    \begin{tabular}{p{0.95\textwidth}}
    {\bf Prompt}  \\
You are provided with a scene description containing multiple rooms. Each room includes a list of objects along with their positions in the room, represented by bounding boxes. Each object's bounding box is defined by a 3D coordinate in the format: <object\_name>(num): [x min, y min, z min],[x max, y max, z max] with units in meters (defining the left-bottom and right-top corners). Your task is to generate an object caption for each room in the form of a coherent, descriptive paragraph that conveys the 3D spatial arrangement and relative positions of all objects within that room.

Then, you will receive the object descriptions and caption for the current 3D room you are in. You will also be provided with the previous rooms' captions as well. 
Your task is to generate new captions covering the summarization of the common features across all rooms based on your current room and important difference based on your current room. 
The reasons of generating the new caption is to help the agent to remind of what are in previous rooms memories can help the agent in this current room. The past objects and observations should be related to current room by examining the summarization of common things and differences. 

\\
    
    For clarity, consider these examples: \textcolor[rgb]{0,0.7,0}{ \{In-context examples\} } \\
------------------- \\
    Here is the scene information: 
\textcolor[rgb]{0.8,0,0}{\{Input scene information\}} \\
    Here is current room you are in and previous rooms you went: 
\textcolor[rgb]{0,0,0.8}{\{Input agent's location\}} \\
\\

    \end{tabular}
\end{tcolorbox}
\caption{Prompt template for generate QA data. \textcolor[rgb]{0,0.7,0}{ \{In-context examples\} } are in-context examples. \textcolor[rgb]{0.8,0,0}{\{Input scene information\}} are scene, room and object semantics along with their bounding boxes. \textcolor[rgb]{0,0,0.8}{\{Input agent's location\}} is the location for current room in the scene and the past explored rooms.}
    \label{tab:generate caption}
\end{minipage}
\end{table*}

\begin{table*}[h!]\centering
\begin{minipage}{0.95\textwidth}   
\centering
\begin{tcolorbox} 
    \centering
    \small
    \begin{tabular}{p{0.95\textwidth}}
   { {\bf System message} } \\

Please act as an impartial judge and evaluate the quality of the response provided by an
AI assistant to the user question. Your evaluation should consider correctness and
helpfulness. You will be given a reference answer and the assistant's answer. You
evaluation should focus on the assistant's answer to the second question. Begin your
evaluation by comparing the assistant's answer with the reference answer. Identify and
correct any mistakes. Be as objective as possible. After providing your explanation, you
must rate the response on a scale of 1 to 10 by strictly following this format:
"[[rating]]", for example: "Rating: [[5]]". \\
    \midrule
    {\bf Prompt}  \\
   <|The Start of Reference Answer|>
   
\#\#\# User:

\text{question\_1}

\#\#\#  Reference answer:

\text{ref\_answer\_1}

\#\#\#  User:

\text{question\_2}

\#\#\#  Reference answer:

\text{ref\_answer\_2}

<|The End of Reference Answer|>

<|The Start of Assistant A's Conversation with User|>

\#\#\#  User:

\text{question\_1}

\#\#\# Assistant A:

\text{answer\_1}

\#\#\# User:

\text{question\_2}

\#\#\#  Assistant A:

\text{answer\_2}

<|The End of Assistant A's Conversation with User|>
\\

    \end{tabular}
\end{tcolorbox}
\caption{Prompt template for open-ended QA evaluation following standard LLM-as-judge protocol.}
    \label{tab: llm as judge}
\end{minipage}
\end{table*}

\section{Data Validation}

\label{appendix: data val}

\subsection{Trajectory Validation}
\label{appendix: traj validation}
We implement a trajectory simulation pipeline driven by the commands listed in Table~\ref{tab:generate trajectory}. For each command, the simulator records the agent’s current room and the full set of objects it is holding, then updates the set of objects in each room to reflect pick-up and put-down actions. A pick-up removes the specified object (along with any nested items) from the room the agent occupies and adds it to the agent’s hand; a put-down removes the object from the agent’s hand and places it into the designated room. The pipeline validates each command based on these criteria: (1) the agent’s location; (2) the referenced object and (3) the correctness of pick-up and put-down actions. For location validation, a command is marked as invalid if the agent attempts to pick up an object from a room that does not match its current room, or tries to drop an object into a room other than the one it currently occupies. Additionally, if the agent tries to visit a room that does not exist in the scene, or attempts to enter a new room when all rooms have already been explored, the trajectory is also considered invalid. For object validation, a pick-up command is invalid if the target object does not exist in the current room, and a put-down command is invalid if the agent is not currently holding the specified object. For pick-up and put-down validation, the agent is allowed to hold only one object at a time. A command is considered invalid if the agent attempts to pick up an object while already holding one, or tries to put down an object when its hand is empty. Finally, after all commands have been executed, if the trajectory ends with the agent still holding an object that was never put down, the entire trajectory is marked as invalid.

\subsection{Human Validation}
\label{appendix: human validation}

\begin{figure}[ht]
  \centering
\includegraphics[trim=0cm 5.8cm 14.1cm 0cm, clip, width=\linewidth]{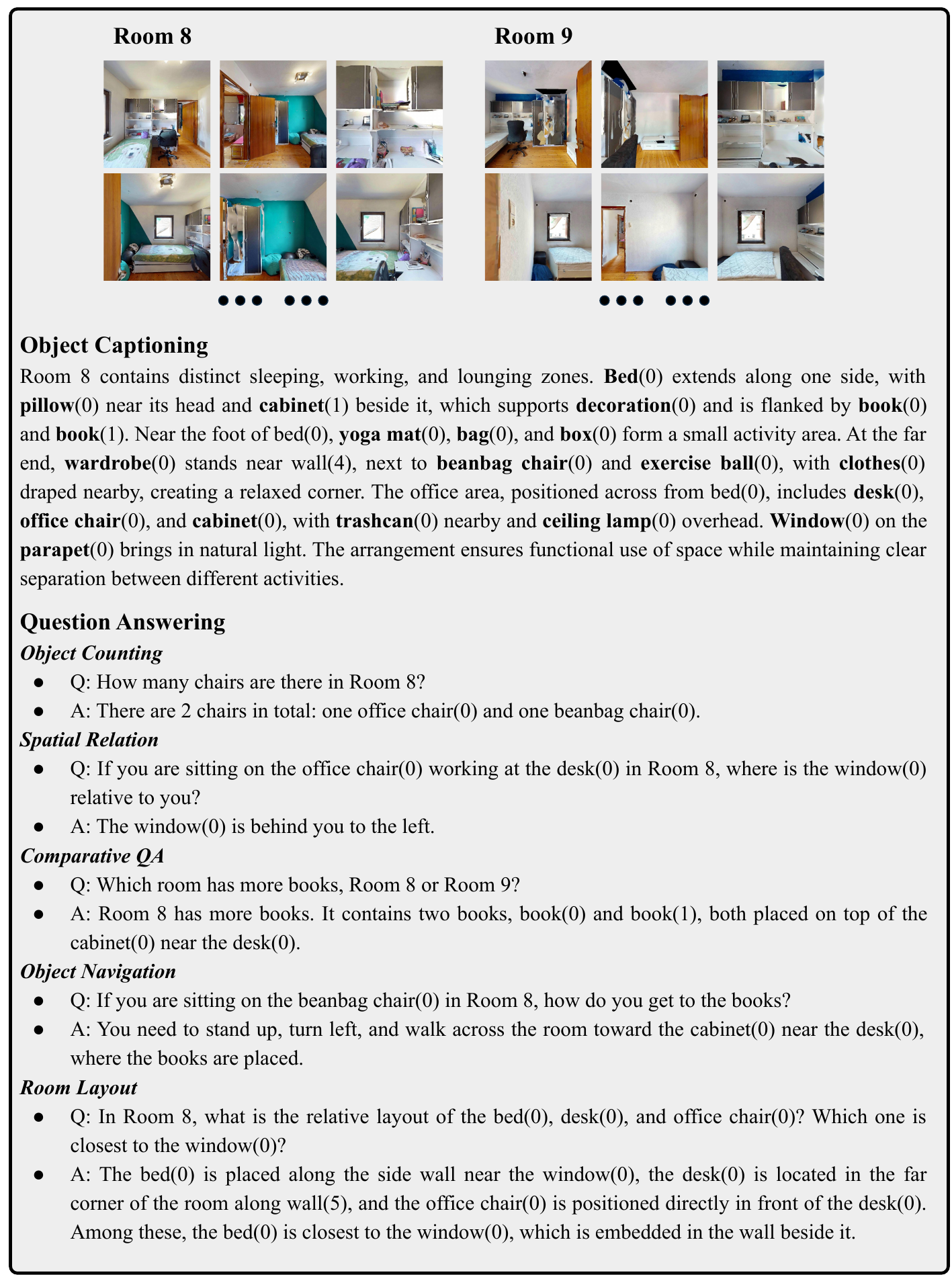}
  \caption{Example of human annotators manually check the data quality on QA and captioning tasks through multiple rendered multi-view images from each room. 
}
 \label{fig:human ann}
 \vspace{-3mm}
\end{figure}

As mentioned in $\S$\ref{sec: data curation} After automatic trajectory validation, we further conduct human validation, in which four student experts in the field manually inspect each benchmark example. We render multi-view images of the entire scene using the simulator and verify whether the benchmark annotations accurately correspond to the simulated environment as illustrated in Figure~\ref{fig:human ann}. 

\section{Evaluation Setup Details}

\label{appendix: eval setup}

\paragraph{3D-LLM}

Similar to the 3D-LLM work~\citep{3dllm}, we use their direct reconstruction method to extract the 3D features from each scene in our training data. To process our long-term memory data, which requires multi-scene input across each task, we feed each room in the task through the 3D-LLm Q-Former head independently to get separate 32-token dense representation of each room with per-room 3d positional embeddings injected into the features. Then we concatenate the representations before feeding the input into the frozen t5-flanxl~\citep{chung2022scaling} backbone like the original work. 
% In our task, the model is not required to understand the positional relations of objects between different rooms in the scene, so we did not add scene-wide positional embeddings.

The 3D-LLM model also included learned location tokens used to describe certain locations within each room in the scene. To fit 3D-LLM to our task data, we substitute the location tokens with our specific interaction tokens (eg. <GO TO ROOM> used by all models in our experiments) and train the model to learn the new tokens to stay consistent with our higher level interaction used across our training data. Analysis of the 3D-LLM model evaluation output, indicated the primary struggle for the model was retaining long term memory of semantic observations in the scene, so we prioritized aligning 3D-LLM with the high level long-term memory representation in our data over low level spatial understanding of the scene.
% \todo{possibly rephrase this to highlight the difference between their goal of spatial understanding and our goal of long term memory}

Our longer task data input also required truncation to fit within the 512 token context length of 3D-LLM's t5-flanxl backbone. We retain the task description and move the question to the beginning of the prompt for the QA data to ensure the model still receives the information necessary to understand its tasks. The longer trajectory of past events is then the only information that gets truncated before fed into the t5 encoder.

For finetuning on our data, we use the hyperparameters provided by 3D-LLM and finetune until model loss stops decreasing. Due to compute limitations, we trained on captioning task for 15 epochs, question-answering task for 20 epochs, and allocated most of the compute time on the embodied task, which we trained on for 75 epochs.

\paragraph{3D-Mem}

We benchmark 3D-Mem~\citep{yang20243dmem3dscenememory} on the question-answering and captioning splits of \benchmark. 3D-Mem is a snapshot-based 3D memory architecture originally developed for embodied exploration and reasoning; it keeps two complementary stores—memory snapshots, a compact set of multi-view RGB-D frames with per-object bounding boxes summarizing the areas the agent has inspected, and frontier snapshots, boundary views that suggest where useful new information may be found next. In its native setting the agent navigates an unfamiliar scene by selecting the frontier view most likely to advance its task and then answers visual questions using the most relevant memory snapshots. Because our evaluation focuses on post-exploration reasoning rather than active exploration, we disable the frontier component and retain only the memory snapshots. For these two tasks, the system will capture memory snapshots in each room from the room center, and finish the QA and captioning base on the memory snapshots of all the explored rooms.

\section{Qualitative Examples}
\label{appendix: qualitative}

\begin{figure}[t]
  \centering
\includegraphics[trim=0cm 10cm 11.3cm 0cm, clip, width=\linewidth]{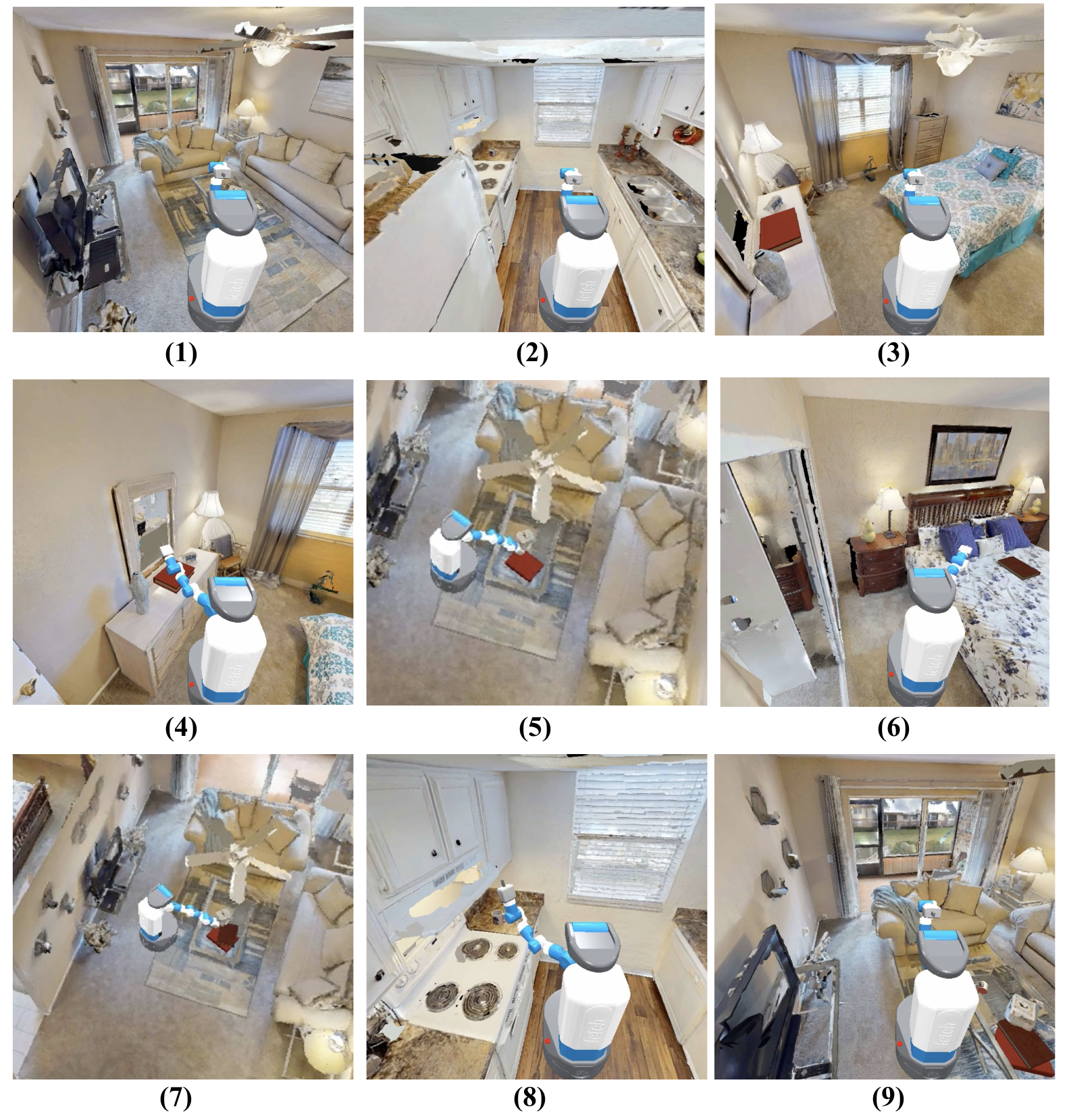}
  % \caption{Qualitative example of \model. The task instruction is: \textit{Prepare a cozy reading nook in the living room with two books and tea}. In the image (1) and (2), the agent is randomly exploring the room and forming an initial memory of the scene. After receiving the task instruction, the agent recalls its memory and navigates to the bedroom to pick up the book on the cabinet as shown in image (3) and (4). Then it returns to the living room and puts it down on the table in front of the sofa as shown in image (5). The agent can't recall any books in its memory and starts exploring new rooms. Then it finds the book on the bed and pick it up as shown in image (6). It returns to the living room and stack the book on top of the other book in image (7). Finally, it recalls its memory and navigates to the kitchen to pick up the tea cup as shown in image (8) and put it on the table in living room as in image (9). The task is successfully complete. }
  \caption{
Qualitative example of \model. The task instruction is: \textit{Prepare a cozy reading nook in the living room with two books and a teacup}. 
In images (1) and (2), the agent explores the environment randomly, forming an initial memory of the scene. After receiving the task instruction, it recalls its memory and navigates to the bedroom to pick up a book from the cabinet, as shown in images (3) and (4). The agent then returns to the living room and places the book on the table in front of the sofa (image 5). 
Unable to recall any additional books, the agent resumes exploration and finds a second book on the bed, which it picks up (image 6) and stacks on top of the first book (image 7). Finally, the agent recalls seeing a teacup in the kitchen, navigates to retrieve it (image 8), and places it on the table in the living room (image 9). The task is successfully completed.
}

 \label{fig:qualitative}
 \vspace{-3mm}
\end{figure}

We provide qualitative examples as shown in Figure~\ref{fig:qualitative}. It demonstrates that \model can maintain a long-term memory and perform complex tasks in the embodied environments. More examples can be found in the \textbf{supplementary materials}. 

\end{document}